\documentclass[twoside,11pt]{article}

\usepackage{arxiv}

\usepackage[latin9]{inputenc}
\usepackage{float}
\usepackage{amsmath}
\usepackage{amssymb}
\usepackage{graphicx}
\usepackage{caption}
\usepackage{subcaption}
\providecommand{\tabularnewline}{\\}
\floatstyle{ruled}
\newfloat{algorithm}{tbp}{loa}
\providecommand{\algorithmname}{Algorithm}
\floatname{algorithm}{\protect\algorithmname}
\DeclareSymbolFont{symbolsC}{U}{txsyc}{m}{n}
\DeclareMathSymbol{\notniFromTxfonts}{\mathrel}{symbolsC}{61}
\usepackage{authblk}
\makeatother
\usepackage{lipsum}

\begin{document}
\title{Controlling for sparsity in sparse factor analysis models: adaptive latent feature sharing for piecewise linear dimensionality reduction}

\author[1]{Adam Farooq}
\author[1*]{Yordan P. Raykov}
\author[2]{Petar Raykov}
\author[3,4]{Max A. Little}

\affil[1]{Department of Mathematics, Aston University, Birmingham, UK}
\affil[2]{School of Psychology, University of Sussex, Brighton, UK}
\affil[3]{Department of Computer Science, University of Birmingham, Birmingham, UK}
\affil[4]{Media Lab, Massachusetts Institute of Technology, Cambridge, MA 02139, USA}


\maketitle
\begin{abstract}

Ubiquitous linear Gaussian exploratory tools such as principal component analysis (PCA) and factor analysis (FA) remain widely used for exploratory analysis, pre-processing, data visualization and related tasks. Due to their rigid assumptions, for very high dimensional problems, they have been replaced by robust, sparse extensions or more flexible discrete-continuous latent feature models. Discrete-continuous latent feature models specify a dictionary of features dependent on subsets of the data and then infer the likelihood of each data point to share any of these features. This most often uses a Beta-Bernoulli model for the feature allocation process which assumes across the features are independent with shared feature dictionary. An undesired consequence of this formalism is that the feature dictionary becomes dominated by the most common, rather than most useful, features. In this work we propose a general alternative feature allocation approach that allows for natural control over the number of features used to express each point, in addition to the control over the whole feature dictionary. This new approach is based on using discrete distribution models without replacement which can adapt to capture both common and rare features. This new framework is used to derive a novel adaptive variant of factor analysis (aFA), as well as, an adaptive probabilistic principal component analysis (aPPCA) capable of flexible structure discovery and dimensionality reduction in a wide variety of scenarios. We derive both standard Gibbs sampler as well as expectation-maximization inference algorithms that converges orders of magnitude faster, to a reasonable point estimate solution. The utility of the proposed aPPCA and aFA models is demonstrated on standard tasks such as feature learning, data visualization and data whitening. We show that aPPCA and aFA can infer interpretable high level features for raw MNIST or COLI-20 images, or when applied to analyse autoencoder features. We also demonstrate that replacing common PCA pre-processing pipelines in the analysis of functional magnetic resonance imaging (fMRI) data with aPPCA leads to more robust and better localized blind source separation of neural activity.

\end{abstract}


\section{Introduction \label{Sec:Introduction}}
\let\thefootnote\relax\footnotetext{*corresponding author}
Latent feature models provide principled and interpretable means for structure decomposition by leveraging specified relationships in the observed data. They are complementary to flexible continuous latent variable models (LVMs) or black-box autoencoder approaches which do not explicitly handle discreteness in the latent space and in fact can often be used in conjunction. A widely occurring application of latent feature models has been as building block for Bayesian \textit{sparse factor analysis}
models \cite{bernardo2003bayesian, knowles2007infinite, Paisley2009, bhattacharya2011sparse, gao2013latent} which are fundamental tools for dimensionality reduction and latent structure discovery in high dimensional data. As a consequence of this, feature allocation priors are designed to induce sparsity, but behave poorly in the presence of even a limited number of dense factors; a problem which can limit their use in the ``big data" problems for which they are intended.

If we are not explicitly interested in capturing discrete allocation of shared factors, alternative Bayesian factor analysis priors have been proposed \cite{engelhardt2010analysis,bhattacharya2011sparse,gao2013latent,legramanti2020bayesian}. In this work, we propose an explicit feature allocation model which can be used to capture both sparse and dense allocation distributions using a discrete urn model without replacement. We propose a flexible set of latent feature linear models, which adopt the \textit{multivariate hypergeometric distribution} as feature allocation process.





Linear dimensionality reduction methods are a mainstay of high-dimensional data analysis, due to their simple geometric interpretation and attractive computational properties. In linear Gaussian LVMs we assume the following generative model for the data:

\begin{equation}
\mathbf{y}=\mathbf{W}\mathbf{x}\:+\boldsymbol{\mu}\:+\:\boldsymbol{\epsilon}\label{eq:Linear-Gaussian-2}
\end{equation}
where the observed data is $\mathbf{y} \in \mathbb{R}^{D}$; $\mathbf{W} \in \mathbb{R}^{D\times K}$ is a transformation matrix the columns of which are commonly referred to as \textit{principal components} or \textit{factors}; $\mathbf{x} \in \mathbb{R}^{K}$ are unknown multivariate Gaussian latent variables, also referred to as factor \textit{loadings}; $\boldsymbol{\mu}\in\mathbb{R}^{D}$ is
a mean (offset) vector and $\boldsymbol{\epsilon}$ describes the model noise, typically Gaussian. Depending on the assumptions we impose on $\mathbf{x}$, $\mathbf{W}$ and $\boldsymbol{\epsilon}$, we can obtain various, widely-used techniques: 
\begin{itemize}
\item The ubiquitous \textit{principal component analysis} (PCA) \cite{pearson1901liiiPCA}
can be derived from Equation \eqref{eq:Linear-Gaussian-2} and further making
the assumptions that $\boldsymbol{\mu}=0$, vectors of $\mathbf{W}$
are orthogonal and the variance of the isotropic noise is 0,
i.e. assume $\boldsymbol{\epsilon}\sim\mathcal{N}\left(\boldsymbol{0},\sigma^{2}\mathbf{I}_{D}\right)$
and $\sigma\to0$ (known as the small variance asymptotic (SVA) assumption). 
\item If we avoid SVA of PCA,
but still assume $\mathbf{W}$ has orthogonal vectors and Gaussian
noise $\boldsymbol{\epsilon}\sim\mathcal{N}\left(\boldsymbol{0},\sigma^{2}\mathbf{I}_{D}\right)$,
we recover \textit{probabilistic PCA} (PPCA) \cite{PPCApaper}. 
\item In the case where we omit the orthogonality assumption on $\mathbf{W}$
and assume more flexible elliptical noise $\boldsymbol{\epsilon}\sim\mathcal{N}\left(\boldsymbol{0},\textrm{diag}\left(\boldsymbol{\sigma}\right)\right)$
with $\boldsymbol{\sigma}=\left(\sigma_{1},\dots,\sigma_{D}\right)$,
we obtain the classic \textit{factor analysis} (FA) \cite{harman1960modernFA}. 
\item Variants of \emph{independent component analysis} \cite{comon1994independentICA}
can be obtained by assuming flexible elliptical noise $\epsilon\sim\mathcal{N}\left(\boldsymbol{0},\textrm{diag}\left(\boldsymbol{\sigma}\right)\right)$
with $\boldsymbol{\sigma}=\left(\sigma_{1},\dots,\sigma_{D}\right)$,
but also assuming a non-Gaussian distribution model for the latent
variables $\mathbf{x}\in\mathbb{R}^{K}$; for example the multivariate Laplace distribution \cite{knowles2007infinite}. 
\end{itemize}

A widely accepted challenge shared by all of these linear Gaussian techniques is that the columns of $\mathbf{W}$ (i.e. the principal components (PCs) or factors) are a linear combination of \textit{all} the original variables. This problem also persists for more flexible continuous latent variable models \cite{dai2015spike} and often makes it difficult to interpret the results. To handle these issues, there has been a plethora of prior work on developing \textit{sparse} PCA \cite{Zou2006sparsePCA} and \textit{sparse} FA models \cite{bernardo2003bayesian}. \cite{Zou2006sparsePCA} places a least absolute shrinkage and selection operator (LASSO) regularization on columns of $\mathbf{x}$, which, compared to simple thresholding, leads to more interpretable components. Similar models have been achieved with the fully Bayesian approach of relevance determination priors \cite{campbell2017probabilistic}. \cite{bernardo2003bayesian} have further suggested placing a two-component mixture model over the loadings $\mathbf{x}$ that allow switching factors from $\mathbf{W}$ to be ``switched" on or off, imposing natural dimensionality reduction. In this scenario, the probabilities of factors having non-zero loadings are independent across all points. \cite{bhattacharya2011sparse} proposed a sparse Bayesian \textit{infinite} factor model which assumes a multiplicative gamma process prior on the loading vectors. This model allows natural inference of the number of latent sparse factors, however it does assume coupling between the portion of explained variance with factor loading sparsity (i.e. sparsity of $\mathbf{x}$). \cite{gao2013latent} aimed to address this issue and proposed an alternative factor analysis setup where a flexible three parameter Beta prior is used on the factor loadings to induce element-specific, factor-specific and global shrinkage. \cite{gao2013latent} also uses a 2-component mixture to cluster each factor as sparse or dense. \cite{durante2017note} provides a great intoduction to some of the issues involed with multiplicative gamma process priors and recently, an additional flexible sparse nonparametric factor analysis prior was proposed in \cite{legramanti2020bayesian}. However, a motivating example in many of these factor models has been the $N<<D$ problem setup with emphasis on learning identifiable factor loadings. 

\begin{figure}
\includegraphics[scale=0.65]{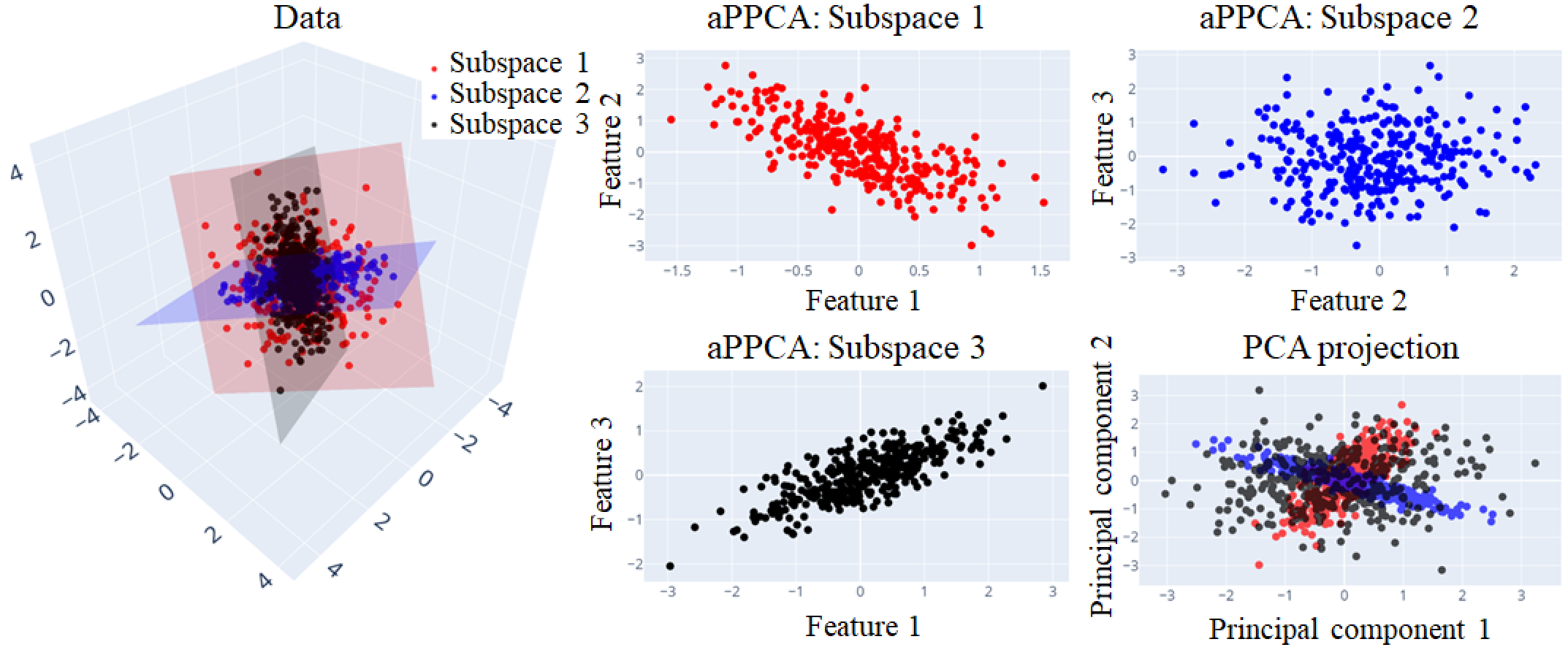}
\centering{}\caption{Illustration of latent feature aPPCA model used for decomposition and dimensionality reduction, plotted against conventional PCA. The left 3-D plot displays synthetic data which lies approximately in one of three separate linear 2-D subspaces which are spanned by different combinations of three orthogonal 1-D PCs. The right subplots display the inferred 2-D projections onto the identified subspaces using aPPCA and PCA.
 \label{fig:APPCA_illustration}} 
\end{figure}

In this work, we are interested in explicitly modelling combinations of different subsets of factors, attempting to capture inherent clustering in the latent space. In such cases, explicit modelling of partitions in the high dimensional input space can be achieved via augmenting the latent space with the addition of discrete latent variables. In \textit{latent feature} models, we denote these latent variables with binary vectors $\mathbf{z}\in\mathbb{R}^K$ which indicate all the features associated with that point. This approach allows the capture of flexible clustering, and also can account for overlapping factors and mixed group membership (see Figure \ref{fig:APPCA_illustration}). This is in contrast to \textit{latent class} LVMs which are designed for subspace clustering \cite{parsons2004subspace, vidal2011subspace} or mixtures of factor analyzers \cite{ghahramani1996algorithm, ghahramani2000variational, campbell2017probabilistic}. The challenge is in designing a sufficiently flexible and intuitive model of the latent feature space. Several nonparametric FA models have addressed this using the \textit{Beta processes} \cite{Paisley2009}, or their marginal \textit{Indian buffet processes} \cite{knowles2007infinite,rai2009infinite} (IBP) which have infinite capacity and can be used to infer the feature space dimensionality (i.e. number of features). However, the IBP imposes some explicit sparsity constraints on the feature allocation distribution which can lead to producing non-interpretable, spurious features and overestimation of the underlying number of features \cite{gao2013latent,Elvira2017}. 

The multivariate hypergeometric model we propose here allows for intuitive control over the sparsity of the feature allocation matrix. We show that the parameters of the hypergeometric prior allow for control over the expected sharing, while the IBP assumes log-linear growth of the number of factors \cite{di2020} and decaying factor representation \cite{Teh2009indian}.
The proposed model is parametric since it fixes the number of unique features instantiated, but at the same time has a different parameter controlling the number of unique features used to represent each data point. This is critical since it allows us to naturally separate (1) features which explain large variance percentage for a small subset of the data from (2) spurious features which explain small variance percentage for a potentially larger subset of the data.
This formulation is natural in the context of data visualization and dimensionality reduction, where natural constraints on the feature representation for each data point occur - in visualization, normally, points are reduced to two or three dimensions; in dimensionality reduction, we model each point with $K << D$ dimensions.

\section{Preliminaries \label{SecPreliminaries}}
\subsection{Latent feature factor analysis models \label{subsec:Infinite-sparse-factor}}
\begin{figure}[t]
\includegraphics[scale=0.3]{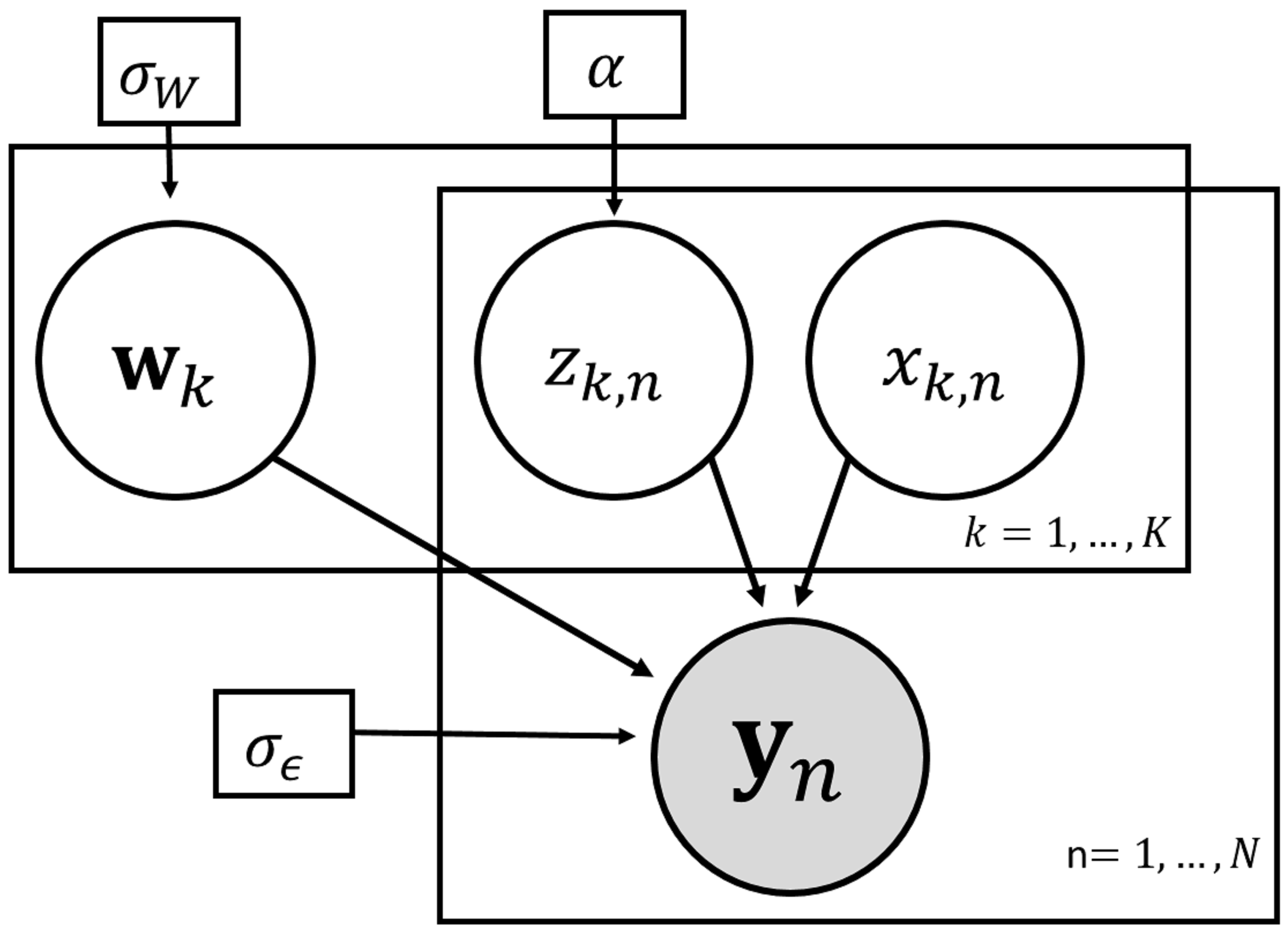}
\centering{}\caption{Graphical model for generic Bayesian latent feature models of which the proposed \textit{adaptive factor analysis} (aFA) and the \textit{adaptive probabilistic principal component analysis} (aPPCA) models are particular examples. If we assume $K\to\infty$ we recover Bayesian nonparametric models such as isFA and iICA. \label{fig:GraphicalModel}}

\end{figure}

In latent feature linear Gaussian LVMs, we augment the model from Equation \eqref{eq:Linear-Gaussian-2} and write the following construction in matrix notation for $N$, $D$-dimensional
observations:

\begin{align}
\mathbf{Y} & =\mathbf{W}(\mathbf{X}\odot\mathbf{Z})+\mathbf{E}\label{eq:isFA}
\end{align}
where $\mathbf{Y} = \left[\mathbf{y}_1,\dots,\mathbf{y}_N\right]$ is the observation matrix,
\textbf{$\mathbf{W}$} is a $\left(D\times K\right)$ factor 
(or mixing) matrix, $\mathbf{Z} = \left[\mathbf{z}_1,\dots,\mathbf{z}_N\right]^T$ is a binary indicator matrix selecting which of $K$ hidden sources are
active, $\odot$ denotes the \textit{Hadamard product}, also known as the element-wise or \textit{Schur product}, $\mathbf{E} = \left[\mathbf{\epsilon}_1,\dots,\mathbf{\epsilon}_N\right]$ is a noise matrix consisting of $N$ independent and identically distributed $D$-dimensional zero-mean vectors drawn from $\mathcal{N}\left(\mathbf{0},\sigma\mathbf{I}_{D}\right)$; finally $\mathbf{X} = \left[\mathbf{x}_1,\dots,\mathbf{x}_N\right]^T$ are the latent variables where each point $x_{k,n}$ is assumed Gaussian for FA and PCA models, and Laplace distributed for Bayesian independent component analysis models. The graphical model is depicted in Figure \ref{fig:GraphicalModel}.

\subsection{Inference}
The joint likelihood of the model (see Figure \ref{fig:GraphicalModel}),  can be written as: 
\begin{align}
\begin{split}
\mathrm{P}\left(\mathbf{Y},\mathbf{W},\mathbf{X},\mathbf{Z}\vert\boldsymbol{\theta}\right) & =\prod_{n=1}^{N}\left(\mathrm{P}\left(\mathbf{y}_{n}\vert\mathbf{W},\mathbf{x}_{n},\mathbf{z}_{n},\sigma\right)\prod_{k=1}^{K}\mathrm{P}\left(x_{k,n}\right)\mathrm{P}\left(z_{k,n}\vert\alpha\right)\right)\\
 & \times\prod_{k=1}^{K}\mathrm{P}\left(\mathbf{w}_{k}\vert\sigma_{W}\right)
\end{split}
\label{Eq:jll_isFA}
\end{align}
where we use $\boldsymbol{\theta}$ to denote jointly the hyperparameters. For the \textit{infinite sparse} FA (isFA) model \cite{knowles2007infinite}, we assume a Gaussian prior on the factor matrix $\mathbf{W}$ and IBP prior on $\mathbf{Z}$ which results in $\boldsymbol{\theta} = \left\{ \alpha,\sigma,\sigma_{W}\right\}$ where $\alpha$ is the concentration parameter for the IBP, $\sigma^{2}$ is the variance of the observed data and $\sigma_{W}^{2}$ is the variance of the factors. We will only briefly summarize a straighforward Gibbs sampler for this isFA model. \cite{Paisley2009} proposed a scalable variational inference algorithm for estimating this model.

The posterior distribution over the latent variables $x_{k,n}$ for which its respective $z_{k,n}=1$ is sampled from a Gaussian:

\begin{equation}
\mathrm{P}\left(x_{k,n}\vert\ldots\right)=\mathcal{N}\left(x_{k,n}\Bigg|\text{\ensuremath{\frac{\mathbf{w}_{k}^{T}\boldsymbol{\epsilon}_{-k,n}}{\sigma^{2}+\mathbf{w}_{k}^{T}\mathbf{w}},\frac{\sigma^{2}}{\sigma^{2}+\mathbf{w}_{k}^{T}\mathbf{w}}}}\right)
\end{equation}
where we have omitted all the variables upon which $x_{k,n}$ depends, $\mathbf{w}_{k}$
is the $k$-th column of the matrix $\mathbf{W}$ and $\boldsymbol{\epsilon}_{-k,n}$
is $\left(\mathbf{y}_{n}-\mathbf{W}(\mathbf{x}_{n}\odot\mathbf{z}_{n})\right)$
with $z_{k,n}=0$, or the noise associated with $n$-th point and $k$-th feature.

The posterior distribution over the $k$-th factor loading $\mathbf{w}_{k}$
is a $D$-dimensional multivariate Gaussian: 

\begin{equation}
\mathrm{P}\left(\mathbf{w}_{k}\vert\ldots\right)=\mathcal{N}\left(\mathbf{w}_{k}\Bigg|\text{\ensuremath{\frac{\sigma_{W}^{2}}{\mathbf{x}_{k}\mathbf{x}_{k}^{T}\sigma_{W}^{2}+\sigma_{\epsilon}^{2}}\mathbf{E}_{-k}\mathbf{x}_{k}^{T},\left(\frac{\mathbf{x}_{k}\mathbf{x}_{k}^{T}}{\sigma_{\epsilon}^{2}}+\frac{1}{\sigma_{W}^{2}}\right)\mathbf{I}_{D}}}\right)
\end{equation}
where $\mathbf{x}_{k}$ is the $k$-th column of the matrix $\mathbf{X}$
and $\mathbf{E}_{-k}$ is $\left(\mathbf{Y}-\mathbf{W}(\mathbf{X}\odot\mathbf{Z})\right)$
with $\mathbf{w}_{k}=\mathbf{0}$.

The matrix $\mathbf{Z}$ is sampled in two steps: the first
involves sampling existing features and the second, sampling new features.
The latent variables $x_{k,n}$ are marginalized out since the collapsed Gibbs sampler can lead to faster convergence
\cite{van2008partially_Gibbs}; the marginal distribution is available
in closed form as the Gaussian prior over the hidden sources is conjugate
to the Gaussian likelihood over the observed data. The existing features
$z_{k,n}$ can be sampled directly using the Bernoulli posterior:

\begin{align}
\begin{split}
    \mathrm{P}\left({z}_{k,n}\vert\ldots\right)\\
    =\mathrm{Bernoulli}\left(\frac{\mathrm{P}\left(\mathbf{y}_{n}\vert z_{k,n} = 1 \right)\mathrm{P}\left(z_{k,n} = 1 \vert \mathbf{z}_{k,-n}\right)}{\mathrm{P}\left(\mathbf{y}_{n}\vert z_{k,n} = 1\right)\mathrm{P}\left(z_{k,n} = 1 \vert \mathbf{z}_{k,-n}\right) + \mathrm{P}\left(\mathbf{y}_{n}\vert z_{k,n} = 0\right)\mathrm{P}\left(z_{k,n} = 0 \vert \mathbf{z}_{k,-n}\right)}\right)
\end{split}
\end{align}

In the described setup, the posterior for new features is not available in closed form,
but it can be approximated using a Metropolis-Hastings step. For each
observation, adding $\kappa$ number of new features and their corresponding
parameters (columns of matrix $\mathbf{W}$) are jointly proposed
and accepted with probability proportional to likelihood improvement
brought about by these new features.

\subsection{Sparsity in Beta-Bernoulli models}\label{sec:Sparsity}

If each component $z_{k,n}$ from the binary vectors $\mathbf{z}_{n}$
is independently drawn from a Bernoulli distribution with the $K$
mixing parameters $\left\{ p_{k}\right\} _{k=1,\ldots,K}$, each independently
drawn from a Beta distribution, then as the number of latent features
$K\to\infty$, one can show that the conjugate prior over the matrix
$\mathbf{Z}^{T}=\left[\mathbf{z}_{1},\ldots,\mathbf{z}_{N}\right]^{T}$
is the \textit{Beta process} \cite{hjort1990nonparametric}. The mixing parameters can be integrated out in order to work with the simpler IBP marginal process. Under the IBP prior,
the indicator matrix $\mathbf{Z}$ is $\left(K\times N\right)$-dimensional
with $K$ being the unknown number of represented features in the
observed data which is assumed to increase with $N$. The expected
number of features $\overline{K}$ follows a Poisson distribution
with mean $\alpha\sum_{n=1}^{N}\frac{1}{N}$; for large $N$, $\overline{K}\approx\alpha\ln\left(N\right)$.
The prior for the matrix $\mathbf{Z}^{T}$ under the IBP is:

\begin{equation}
\begin{array}{c}
\mathrm{P}\left(\mathrm{\mathbf{Z}}^{T}\vert\alpha\right)\propto\exp\left(-\alpha H_{N}\right)\alpha^{K}\left(\prod_{k=1}^{K}\frac{\left(m_{k}-1\right)!\left(N-m_{k}\right)!}{N!}\right)
\end{array}\label{eq:IBPMatrix}
\end{equation}
where $H_{N}=\sum_{n=1}^{N}\frac{1}{n}$ and $m_{k}=\sum_{n=1}^{N}z_{k,n}$.

The IBP prior enforces sparse $\mathbf{Z}^T$ by placing diminishing probability on the event of having many common features $k$, i.e. features with large $m_k$. It has been observed that the number of observations being active in each feature follows \textit{Zipf's law} \cite{Teh2009indian,zipf1932selected}; this implies that a small number of observations are active in all features; and a large number of observations are only active in a small number of features. This Zipf's law behavior has been observed and proven as $N\rightarrow\infty$ \cite{Teh2009indian}, the distribution which models the number of the features as approximately proportional to the reciprocal of the feature size. The first few principal components (PCs) explain a larger proportion of the variance, and are more likely to be shared by a large number of points. Then, from Zipf's law, most of the remaining PCs are a linear combination of just a few observations. In summary, the IBP prior is appropriate in scenarios where we want to induce a sparse feature allocation process, but falls short in cases where we seek a composition of dense features or a mixture of sparse and dense features.

\section{Feature allocation without replacement: introducing the adaptive factor analysis (aFA) model \label{adaptiveFA}}

In this section we propose a novel latent feature FA model which allows us to explicitly specify both the total number of factors (i.e. features) and the number of unique factors each observation is associated with: we refer to it as adaptive factor analysis (aFA). The aFA model can capture a wider set of allocation modalities and addresses many of the practical problems arising in sparse FA models. Motivated by the prohibitive computational costs involved with training most existing latent feature FA models, we derive a scalable expectation-maximization (EM) algorithm for approximate inference in the aFA model.

\subsection{Feature allocation models without replacement}
To implement aFA, we place  a \textit{multivariate hypergeometric distribution} \cite{chesson1976non} as a prior over all the latent feature indicators $\mathbf{Z}$:

\begin{equation}
\mathrm{P}\left(\mathbf{Z}\vert K,L,m_{1,\ldots,K}=1\right)=\prod_{n=1}^{N}\frac{\binom{1}{z_{1,n}}\binom{1}{z_{2,n}}\times\ldots\times\binom{1}{z_{K,n}}}{\binom{K}{L}}\label{eq:Zprior-1-1}
\end{equation}
for given hyperparameter values of $L$ and $K$, such that $L < K$;  $z_{k,n}\in\left\{ 0,1\right\} $ under the constraint that $\sum_{k=1}^{K}z_{k,n}=L$. The parameter $L$ allows for explicit control over the number of latent factors used to decompose each observation. $K$ denotes the number of unique factors used to represent the data, i.e. the number of columns in $\mathbf{W}$. This implies that each input data point is associated with a different subset of $L$ factors, selected from a total of $K$ unique factors. The parameter $K$ accounts for the global sharing of structure across overlapping groups of data points with common factors; if $K$ is large enough, each point can, in principle, be associated with non-overlapping subsets of $L$ factors, equivalent to mixture of the FAs model. But, as $K$ reduces, more of these factors are constrained to be shared across subsets of the data. $L$ acts much like the number of latent dimensions in traditional linear LVMs, but here $L$ is constrained by $K$. This allows us to interpret $L$ as the \emph{local capacity} of the model and $K$ controls \emph{global} capacity of sharing. If $L=K$, we recover classical FA models, since all features are associated with all observed data points. As $K-L$ increases, more local structure in the data can be represented.

\subsection{Scalable inference for aFA}

The joint likelihood for the proposed model takes the same form as FA from Equation \eqref{Eq:jll_isFA}, but with a different distribution over $\mathbf{Z}$. The parametric nature of the hypergeometric model allows us to derive an efficient EM algorithm for training the aFA model, which can be used both for initialization of a full Gibbs sampler or for rapidly obtaining a (local) maximum-a-posteriori solution for the model. We marginalize over the continuous latent variables $\mathbf{X}$
and at each iteration we compute the expectation of the likelihood
with respect to $\mathbf{X}$: $\mathbb{E}_{\mathbf{X}|\ldots}\left[\mathrm{P}\left(\mathbf{Y},\mathbf{Z},\mathbf{X}\vert\mathbf{W},\sigma,\sigma_{x}\right)\times\mathrm{P}(\mathbf{W})\right]$
where $\mathbb{E}_{\mathbf{X}|\ldots}$ denotes conditional expectation
with respect to $\mathrm{P}\left(\mathbf{X}\vert\mathbf{Y},\mathbf{Z},\mathbf{W},\sigma,\sigma_{x}\right)$
and $\sigma_{x}^{2}$ is the variance over the latent space. The log-likelihood
can be expressed as:
\begin{equation}
\begin{split}
    \mathcal{L}_{N}&= -\sum_{n=1}^{N}\Biggl(\frac{K}{2}\ln\left(\sigma_{x}^{2}\right)+\frac{D}{2}\ln\left(\sigma^{2}\right)\\&+\frac{1}{2\sigma_{x}^{2}}\mathbf{x}_{n}^{\mathrm{T}}\mathbf{x}_{n}+\frac{1}{2\sigma^{2}}\mathbf{y}_{n}^{\mathrm{T}}\mathbf{y}_{n}-\frac{1}{\sigma^{2}}\mathbf{x}_{n}^{\mathrm{T}}\mathbf{A}_{n}^{\mathrm{T}}\mathbf{W}^{\mathrm{T}}\mathbf{y}_{n}+\frac{1}{2\sigma^{2}}\mathbf{x}_{n}^{\mathrm{T}}\mathbf{A}_{n}^{\mathrm{T}}\mathbf{W}^{\mathrm{T}}\mathbf{W}\mathbf{A}_{n}\mathbf{x}_{n}\Biggl)
\end{split}
\end{equation}
where $\mathbf{A}_{n}$ is a $\left(K\times K\right)$ matrix with
the diagonal elements being $\mathbf{z}_{n}$. The expectation of
$\mathbf{x}_{n}$ from above can then be written as:
\begin{equation}
    \mathbb{E}\left[\mathbf{x}_{n}\right]=\left(\sigma_{x}^{-2}\mathbf{I}_{K}+\sigma^{-2}\mathbf{A}_{n}^{\mathrm{T}}\mathbf{W}^{\mathrm{T}}\mathbf{W}\mathbf{A}_{n}\right)^{-1}\left(\sigma^{-2}\mathbf{A}_{n}^{\mathrm{T}}\mathbf{W}^{\mathrm{T}}\mathbf{y}_{n}\right)
\end{equation}
Using $\mathbb{E}\left[\mathbf{x}_{n}\right]$, we can rewrite the marginal log-likelihood after integrating $\mathbf{x}_{n}$: 
\begin{equation}
\begin{split}
\mathcal{L}_{N}&=-\sum_{n=1}^{N}\Biggl(\frac{K}{2}\ln\left(\sigma_{x}^{2}\right)+\frac{D}{2}\ln\left(\sigma^{2}\right)\\&+\frac{1}{2\sigma_{x}^{2}}\mathrm{tr}\left(\mathbb{E}\left[\mathbf{x}_{n}\mathbf{x}_{n}^{\mathrm{T}}\right]\right)+\frac{1}{2\sigma^{2}}\mathbf{y}_{n}^{\mathrm{T}}\mathbf{y}_{n}-\frac{1}{\sigma^{2}}\mathbb{E}\left[\mathbf{x}_{n}\right]^{\mathrm{T}}\mathbf{A}_{n}^{\mathrm{T}}\mathbf{W}^{\mathrm{T}}\mathbf{y}_{n}\\&\quad+\frac{1}{2\sigma^{2}}\mathrm{tr}\left(\mathbf{A}_{n}^{\mathrm{T}}\mathbf{W}^{\mathrm{T}}\mathbf{W}\mathbf{A}_{n}\mathbb{E}\left[\mathbf{x}_{n}\mathbf{x}_{n}^{\mathrm{T}}\right]\right)\Biggl)
\end{split}
\end{equation}

In the EM maximization step, we update the rest of the parameters
and the indicator variables by solving $\frac{\partial\mathcal{L}_{N}}{\partial\mathbf{W}}$,
$\frac{\partial\mathcal{L}_{N}}{\partial\sigma}$, $\frac{\partial\mathcal{L}_{N}}{\partial\sigma_{x}}$
and $\frac{\partial\mathcal{L}_{N}}{\partial\mathbf{z}_{n}}=0$.

Since we are often interested only in a point estimate for the indicator
variables $\mathbf{Z}$, iterative optimization via coordinate descent
can lead to a robust, local MAP estimate i.e. $\mathbf{Z}^{\text{MAP}}$ \cite{wang2011,broderick2013mad,raykov2016simple,raykov2017deterministic}.
The complete EM algorithm for the proposed aFA is summarized in Algorithm
\ref{alg:MEM}. Typically, it converges in only a few iterations and
later we show its MAP decomposition leads to comparable reconstruction error to a Gibbs trained aFA. The EM algorithm for aFA will also lead to lower reconstruction error compared to other well known parametric and nonparametric FA algorithms. 

\begin{algorithm}
\textbf{\small{}Input:}{\small{} $\mathbf{Y},\boldsymbol{\Theta},\mathrm{MaxIter}$}{\small\par}

\textbf{\small{}Initialise:}{\small{} Sample a random $\left(K\times N\right)$
binary matrix $\mathbf{Z}$ and initialize $\left\{ \mathbf{W},\mathbf{X}\right\} $ using PCA}{\small\par}

\textbf{\small{}for}{\small{} $\mathrm{iter}\leftarrow1$ to $\mathrm{MaxIter}$}{\small\par}

{\small{}$\quad$}\textbf{\small{}for}{\small{} $n\leftarrow1$ to
$N$}{\small\par}

{\small{}$\quad$$\quad$Set $\mathcal{I}=\left\{ k:z_{k,n}=1\right\} $}{\small\par}

{\small{}$\quad$$\quad$}\textbf{\small{}for $l\leftarrow1$ }{\small{}to
$L$}{\small\par}

{\small{}$\quad$$\quad$$\quad$Set $z_{\mathcal{I}_{l},n}=0$}{\small\par}

{\small{}$\quad$$\quad$$\quad$Sample $\mathcal{I}_{l}$ using \eqref{eq:Parametric_APPCA_Z}}{\small\par}

{\small{}$\quad$$\quad$$\quad$Set $z_{\mathcal{I}_{l},n}=1$}{\small\par}

{\small{}$\quad$}\textbf{\small{}for}{\small{} $n\leftarrow1$ to
$N$}{\small\par}

{\small{}$\quad$$\quad$Set $\mathbf{x}_{n}=\left(\sigma_{x}^{-2}\mathbf{I}_{K}+\sigma^{-2}\mathbf{A}_{n}\mathbf{W}^{\mathrm{T}}\mathbf{W}\mathbf{A}_{n}\right)^{-1}\left(\sigma^{-2}\mathbf{A}_{n}\mathbf{W}^{\mathrm{T}}\mathbf{y}_{n}\right)$}{\small\par}

{\small{}$\quad$$\quad$Set $\boldsymbol{\Psi}_{n}=\left(\sigma_{x}^{-2}\mathbf{I}_{K}+\sigma^{-2}\mathbf{A}_{n}\mathbf{W}^{\mathrm{T}}\mathbf{W}\mathbf{A}_{n}\right)^{-1}+\mathbf{x}_{n}\mathbf{x}_{n}^{T}$}{\small\par}

{\small{}$\quad$Set $\mathbf{W}=\left(\sum_{n=1}^{N}\mathbf{y}_{n}\left(\mathbf{A}_{n}\mathbf{x}_{n}\right)^{\mathrm{T}}\right)\left(\sum_{n=1}^{N}\mathbf{A}_{n}\boldsymbol{\Psi}_{n}\mathbf{A}_{n}\right)^{-1}$}{\small\par}

{\small{}$\quad$Set $\sigma^{2}=\frac{1}{ND}\sum_{n=1}^{N}\left(\mathbf{y}_{n}^{\mathrm{T}}\mathbf{y}_{n}-2\mathbf{x}_{n}^{T}\mathbf{A}_{n}\mathbf{W}^{\mathrm{T}}\mathbf{y}_{n}+\mathrm{trace}\left(\mathbf{A}_{n}\mathbf{W}^{\mathrm{T}}\mathbf{W}\mathbf{A}_{n}\boldsymbol{\Psi}_{n}\right)\right)$}{\small\par}

{\small{}$\quad$Set $\sigma_{x}^{2}=\frac{1}{NK}\sum_{n=1}^{N}\mathrm{trace}\left(\boldsymbol{\Psi}_{n}\right)$}{\small\par}

\caption{EM algorithm for parametric adaptive factor (aFA) analysis.}
\label{alg:MEM}
\end{algorithm}

\section{Latent feature subspace models\label{sec:The-adaptive-probabilistic PCA}}
Latent feature visualization counterparts have received a lot less attention, despite the popularity of sparse principal component analysis techniques \cite{Zou2006sparsePCA,jolliffe2003}. This is most likely due to the complexity of specifying distributions over orthogonal matrices and the difficulty of performing inference with them. In this section, we extend the Bayesian nonparametric FA model from \cite{knowles2007infinite} to the PPCA setup in which the columns of the transformation matrix $\mathbf{W}$ are orthogonal. We argue that the nonparametric PPCA is likely to suffer from the same limitations, as isFA, in the presence of dense PCs and introduce an efficient \textit{adaptive probabilistic principal component analysis} (aPPCA) framework which uses hypergeometric feature allocations. The proposed aPPCA allows for explicit control over both the number of unique columns $K$ in $\mathbf{W}$, as well as the observation-specific number of active vectors $L$.

Latent feature subspace models can be described as a latent feature approach, in which the latent features are assumed to share orthogonal one-dimensional subspaces,
characterized via the projection vectors $\boldsymbol{\mathrm{w}}_{1},\boldsymbol{\mathrm{w}}_{2},...,\boldsymbol{\mathrm{w}}_{K}$ forming $\mathbf{W}$. If two points $\mathbf{y}_{i}$ and $\mathbf{y}_{j}$
are associated with a projection vector $\boldsymbol{\mathrm{w}}_{k}$, it means that sufficient information about these points can be preserved
by projecting them in the direction specified by $\boldsymbol{\mathrm{w}}_{k}$.

Both the nonparametric and the adaptive PPCA models share the following construction:

\begin{align}
\mathbf{y}_{n} & =\mathbf{W}(\mathbf{x}_{n}\odot\mathbf{z}_{n})+\mathbf{\boldsymbol{\mu}}+\boldsymbol{\epsilon}_{n}\nonumber \\
\mathbf{x}_{n} & \sim\,\mathcal{N}\left(0,\mathbf{I}_{K}\right)\label{eq:Adaptive PCA - parametric-2}\\
\boldsymbol{\epsilon}_{n} & \sim\,\mathcal{N}\left(0,\sigma^{2}\mathbf{I}_{D}\right)\nonumber 
\end{align}
for $n=1,\dots N$, where $\mathbf{y}_{n}\in\mathfrak{\mathbb{R}}^{D}$
is the $D$-dimensional observed data; $\mathbf{x}_{n}\in\mathfrak{\mathbb{R}}^{K}$ is the lower dimensional latent variable;
$\mathbf{W}=\left[\boldsymbol{\mathrm{w}}_{1},\boldsymbol{\mathrm{w}}_{2},...,\boldsymbol{\mathrm{w}}_{K}\right]$
is an unobserved $\left(D\times K\right)$ projection matrix with
$\boldsymbol{\mathrm{w}}_{i}\perp\boldsymbol{\mathrm{w}}_{j}\text{ for all }i\neq j$; $\mathbf{z}_{n}\in\mathfrak{\mathbb{R}}^{K}$ is
a binary vector indicating the active subspaces for
point $n$, $\boldsymbol{\epsilon}_{n}$
is zero-mean Gaussian noise; and without loss of generality we assume
the $D$-dimensional mean vector $\mathbf{\boldsymbol{\mu}}=\frac{1}{N}\sum_{n=1}^{N}\mathbf{y}_{n}$
is zero. Condition on the model parameters and latent variables, for both latent feature subspace models we can write the
likelihood of point $n$ as:

\begin{align}
\mathrm{P}\left(\mathbf{y}_{n}\mid\mathbf{W},\mathbf{x}_{n},\mathbf{z}_{n},\sigma\right) & =\frac{1}{\left(2\pi\sigma^{2}\right)^{\frac{D}{2}}}\exp\left(-\frac{1}{2\sigma^{2}}\left(\mathbf{y}_{n}-\mathbf{W}\left(\mathbf{x}_{n}\odot\mathbf{z}_{n}\right)\right)^{T}\left(\mathbf{y}_{n}-\mathbf{W}\left(\mathbf{x}_{n}\odot\mathbf{z}_{n}\right)\right)\right)\label{eq:APPCAlikelihood-1}
\end{align}

\subsection{Inference in latent feature subspace models}
Computing the posterior distribution of the latent variables $\{\mathbf{X},\mathbf{Z}\}$ and the projection matrix $\mathbf{W}$ is analytically intractable and we have to resort to approximate inference. Unlike for aFA above, the posterior updates of the orthonormal matrix $\mathbf{W}$ do not allow for closed form updates. At the same time numerically optimizing over $\mathbf{W}$ and marginalizing $\mathbf{X}$ leads to slow mixing and an EM scheme leads to poor local solutions for this model. An efficient Markov Chain Monte Carlo (MCMC) scheme \cite{gelman2013bayesian} can be derived which iterates between explicit updates for $\textbf{W}$, $\mathbf{z}_{n}$, $\mathbf{x}_{n}$ and the hyperparameters we wish
to infer, i.e. $\sigma^{2}$ and $\alpha$ (update of $\sigma^{2}$
and $\alpha$ is in given in Appendix \ref{Appendix:DistributionParameters}). Sampling from directional posteriors is prohibitively slow, so we propose a MAP scheme for the updates on $\textbf{W}$. Alternatively we could use an automated MCMC platforms such as STAN \cite{Carpenter2017stan} for the inference, but STAN does not deal well with  discontinuous likelihood models such as aPPCA. This can be addressed using discrete relaxations such as \cite{maddison2016concrete} or numerical solver extensions such as \cite{Nishimura2017discontinuous}. However, such an approach can be justified only for nonlinear intractable extensions of latent feature PPCA, since the Gibbs sampler with closed form updates is substantially more efficient.

The joint data likelihood of both latent feature subspace models we propose takes the
form: 
\begin{align}
\begin{split}
\mathrm{P}\left(\mathbf{Y},\mathbf{W},\mathbf{X},\mathbf{Z}\vert  \sigma,  \alpha \right) & =\prod_{n=1}^{N}\left(\mathrm{P}\left(\mathbf{y}_{n}\vert\mathbf{W},\mathbf{x}_{n},\mathbf{z}_{n},\sigma\right)\prod_{k=1}^{K}\mathrm{P}\left(x_{k,n}\right)\mathrm{P}\left(z_{k,n}\vert\alpha\right)\right)\\
 & \times\mathrm{P}\left(\mathbf{W}\right)
\label{eq:Complete_data_likelihood}
\end{split}
\end{align}
We can check whether the MCMC sampler has converged using standard
tests such as \cite{raftery1992practical} directly on Equation \eqref{eq:Complete_data_likelihood}.
Comparing the Bayesian nonparametric sparse PPCA model and the aPPCA model,
the only difference is in $\mathrm{P}\left(\textbf{Z}\right)$. We
will see that this will affect the posterior update of $\textbf{Z}$,
but the rest of inference algorithm is otherwise identical across
both models.

\subsubsection*{Posterior of $\textbf{W}$}
In order to comply with the orthogonality constraint on $\mathbf{W}$, i.e. $\boldsymbol{\mathrm{w}}_{i}\perp\boldsymbol{\mathrm{w}}_{j}\;\forall i\neq j$, we have to use a distribution with support on the Stiefel manifold (see \cite{tagare2011notesStefiel} for a good introduction).  \cite{Elvira2017} explored exactly this problem in the context of latent feature subspace modelling and proposed using a conjugate \textit{Bingham} prior \cite{Bingham1974} independently on the columns of $\mathbf{W}$ leading to an independent \textit{von Mises-Fisher} posterior over each column where re-scaling is required after each sample to maintain orthogonality. However, empirical trials suggest that this results in very poor mixing. To overcome this issue, we propose joint sampling of the columns of $\mathbf{W}$. We place a uniform prior over the Stiefel manifold on the matrix $\mathbf{W}$ which allows us to work with a  \textit{matrix von Mises-Fisher} \cite{khatri1977mises} posterior:

\begin{equation}
\mathrm{P}\left(\mathbf{W}\vert\mathbf{Y},\mathbf{X},\mathbf{Z},\sigma\right)=\mathrm{_{0}F_{1}^{-1}}\left(\emptyset,\frac{D}{2},\mathbf{A}\mathbf{A}^{T}\right)\exp\left(\mathrm{tr}\left(\mathbf{A}\mathbf{W}\right)\right)
\label{eq:PostW}
\end{equation}
where $\mathbf{A}=\frac{1}{2\sigma^{2}}\left(\mathbf{X}\odot\mathbf{Z}\right)\mathbf{Y}^{T}$
and $\mathrm{_{0}F_{1}^{-1}}\left(\cdot\right)$ is a hypergeometric
function \cite{herz1955bessel}. The normalization term of the matrix
von Mises-Fisher posterior is not available in closed form, hence
it is common to sample from it using rejection sampling. \cite{fallaize2016Bingham}
proposed a Metropolis-Hastings scheme to generate samples from Equation \eqref{eq:PostW},
the resulting posterior of $\mathbf{W}$ converges faster than the
 Bingham-von-Mises-Fisher posterior, but can be further
sped up by numerical optimization methods. Here, we propose updating
the matrix $\mathbf{W}$ by maximizing the posterior from Equation \eqref{eq:PostW}
over the Stiefel manifold, i.e. keeping orthogonality $\boldsymbol{\mathrm{w}}_{i}\perp\boldsymbol{\mathrm{w}}_{j}\;\forall i\neq j$.
An efficient implementation can be achieved using the \textsc{Pymanopt}
toolbox \cite{townsend2016pymanopt}, for optimization over manifolds
with different geometries; this step is outlined in Appendix \ref{Appendix:W-update}.

\subsubsection*{Posterior of $\textbf{X}$}

The posterior distribution over the latent variable $x_{k,n}$, for which
its respective $z_{k,n}=1$, is sampled from a Gaussian:

\begin{equation}
\mathrm{P}\left(x_{k,n}\vert\mathbf{w}_k,\mathbf{y}_n,\mathbf{z}_n \right)=\mathcal{N}\left(x_{k,n}\left|\frac{\textbf{y}^T_n \textbf{w}_k}{\sigma^2 + 1}, \frac{\sigma^2}{\sigma^2 + 1}\right)\right)
\label{eq:A-PPCAPostX}
\end{equation}
where $\mathbf{w}_{k}$ is the $k$th column of the matrix $\mathbf{W}$.
\subsubsection{Bayesian nonparametric sparse PPCA model}\label{sec:BNP A-PPCA}
In the Bayesian nonparametric PPCA, we place an IBP prior over
the indicator matrix $\mathbf{Z}$; this assumes that after a finite
$N$ number of observations only a finite $K$ number of one-dimensional
subspaces are active. This results in the first $K$ rows of $\mathbf{Z}$
having non-zero entries, and the remaining
being all zeros. By design, $K$ cannot exceed the dimension
of the data $D$ and this leads to truncation of the IBP such that
$K$ has a upper limit of $K^{\mathrm{max}};$ where $K\leq K^{\mathrm{max}}\leq D$,
therefore in the Bayesian nonparametric PPCA, \textbf{Z} is a $\left(K^{\mathrm{max}}\times N\right)$
binary matrix, with the sum of the first $K$ rows being non-zero
and the sum of the remaining $K^{\mathrm{max}}-K$ rows being zero.
 We sample the matrix $\mathbf{Z}$
in two stages which include sampling ``existing features'' and``new
features''; in both cases the latent variables $x_{k,n}$ are marginalized
out. The posterior distribution over the existing features $z_{k,n}$
is Bernoulli distributed:

\begin{equation}
\begin{split}\mathrm{P}\left(z_{k,n}\vert\ldots\right)=\\
= & \mathrm{Bernoulli}\left(\frac{\mathrm{P}\left(\mathbf{y}_{n}\vert z_{k,n}=1\right)\mathrm{P}\left(z_{k,n}=1\vert\mathbf{z}_{k,-n}\right)}{\mathrm{P}\left(\mathbf{y}_{n}\vert z_{k,n}=1\right)\mathrm{P}\left(z_{k,n}=1\vert\mathbf{z}_{k,-n}\right)+\mathrm{P}\left(\mathbf{y}_{n}\vert z_{k,n}=0\right)\mathrm{P}\left(z_{k,n}=0\vert\mathbf{z}_{k,-n}\right)}\right)\\
= & \mathrm{Bernoulli}\left(\frac{\frac{m_{k,-n}}{N}\exp\left(\frac{1}{2\sigma^{2}\left(\sigma^{2}+1\right)}\left(\mathbf{y}_{n}^{T}\mathbf{w}_{k}\right)\right)\left(\frac{\sigma^{2}}{\sigma^{2}+1}\right)^{\frac{1}{2}}}{\frac{m_{k,-n}}{N}\exp\left(\frac{1}{2\sigma^{2}\left(\sigma^{2}+1\right)}\left(\mathbf{y}_{n}^{T}\mathbf{w}_{k}\right)\right)\left(\frac{\sigma^{2}}{\sigma^{2}+1}\right)^{\frac{1}{2}}+1}\right)
\end{split}
\label{eq:NonParametricAPPCA_ExistingZ}
\end{equation}
where we omit the dependence on $\mathbf{W}$ and $\sigma^{2}$ and
$m_{k,-n}=\sum_{i\neq n}z_{k,i}$.

Then, we sample $\kappa$ number of new features with $\kappa\sim\text{Poisson}\left(\frac{\alpha}{N}\right)$,
where we maintain $\kappa>0$ or $\kappa+K\leq K^{\mathrm{max}}$.
For observed data point $n$, the posterior distribution over the
new features is: 

\begin{equation}
\mathrm{P}\left(z_{K+j,n}\vert\ldots\right)=\text{Bernoulli}\left(\frac{\exp\left(\frac{1}{2\sigma^{2}\left(\sigma^{2}+1\right)}\sum_{k=K+1}^{K+\kappa}\left(\mathbf{y}_{n}^{T}\mathbf{w}_{k}\right)^{2}\right)\left(\frac{\sigma^{2}}{\sigma^{2}+1}\right)^{\frac{\kappa}{2}}}{\exp\left(\frac{1}{2\sigma^{2}\left(\sigma^{2}+1\right)}\sum_{k=K+1}^{K+\kappa}\left(\mathbf{y}_{n}^{T}\mathbf{w}_{k}\right)^{2}\right)\left(\frac{\sigma^{2}}{\sigma^{2}+1}\right)^{\frac{\kappa}{2}}+1}\right)
\label{eq:NonParametricAPPCA_NewZ}
\end{equation}
for $j=1,\dots,\kappa$ new features.

\begin{algorithm}
\textbf{\small{}Input:}{\small{} $\mathbf{Y},\boldsymbol{\Theta},\mathrm{MaxIter}$,
$K$}{\small\par}

\textbf{\small{}Initialise:}{\small{} Sample a random $\left(K^{max}\times N\right)$
binary matrix $\mathbf{Z}$ and initialize $\mathbf{W}$ using PCA}{\small\par}

\textbf{\small{}for}{\small{} $\mathrm{iter}\leftarrow1$ to $\mathrm{MaxIter}$}{\small\par}

{\small{}$\quad$}\textbf{\small{}for}{\small{} $n\leftarrow1$ to
$N$}{\small\par}

{\small{}$\quad$$\quad$}\textbf{\small{}for}{\small{} $k\leftarrow1$
to $K$}{\small\par}

{\small{}$\quad$$\quad$$\quad$Sample $z_{k,n}$ using \eqref{eq:NonParametricAPPCA_ExistingZ}}{\small\par}

{\small{}$\quad$$\quad$Sample $\kappa\sim\mathrm{Poisson}\left(\frac{\alpha}{N}\right)$}{\small\par}
{\small{}$\quad$$\quad$Accept $\kappa$ new features with probability \eqref{eq:NonParametricAPPCA_NewZ} and update $K$ accordingly}{\small\par}

{\small{}$\quad$}\textbf{\small{}for}{\small{} $n\leftarrow1$ to
$N$}{\small\par}

{\small{}$\quad$$\quad$}\textbf{\small{}for}{\small{} $k\leftarrow1$
to $K$}{\small\par}

{\small{}$\quad$$\quad$$\quad$}\textbf{\small{}if }{\small{}$z_{k,n}=1$}{\small\par}

{\small{}$\quad$$\quad$$\quad$$\quad$Sample $x_{k,n}$ using
\eqref{eq:A-PPCAPostX}}{\small\par}

{\small{}$\quad$Sample $\mathbf{W}$ using \eqref{eq:PostW}}{\small\par}

{\small{}$\quad$Sample $\left\{ \sigma^{2},\alpha\right\}$ from Appendix \ref{Appendix:DistributionParameters}}{\small\par}

\caption{Pseudocode for inference in Bayesian nonparametric PPCA using Gibbs
sampling.}
\label{alg:NonParaGibbs}
\end{algorithm}

\subsubsection{Learning robust subspace features with the adaptive PPCA}

In many common PPCA applications, constraints on the latent feature dimensionality occur naturally.
In data visualization, we are mostly interested in reducing high dimensional data down to two or three dimensions;
in regression problems when PCA is used to remove \textit{multicollinearity} from input features, the output dimensionality is usually fixed to $D$ (the dimensionality of the input). In these scenarios the multivariate hypergeometric model for  $\mathbf{Z}$ allows explicit control over the number of latent subspaces
$L$ used to decompose each single observation. $K$ denotes the
number of unique orthogonal linear subspaces which we will use to
reduce the original data into the lower dimensional space; each input data point can be associated with different
subset of $L$ subspaces, selected from a total of $K$ subspaces.
So, any single point is actually represented by lower dimensional spaces subsets of $\mathbb{R}^{L}$. Note that the orthogonality assumption $\boldsymbol{\mathrm{w}}_{i}\perp\boldsymbol{\mathrm{w}}_{j}\;\forall i\neq j$
for the columns of $\mathbf{W}$ implies that $K\leq D$.

The hypergeometric prior allows updates of $\mathbf{Z}$
across $N$ in parallel, since the number of observed data points
assigned to a latent subspace no longer implies higher probability
of assigning a new data point to that subspace, i.e. no
reinforcement effect. Instead, for each $n=1,\ldots,N$, we sample
$\mathbf{z}_{n}$ by first finding the $L$ observed data indices
$\left\{ l_{1},\ldots,l_{L}\right\} $ for which $\mathbf{z}_{n}$, then for each $l_{i}$, we set $z_{n,l_{i}}=0$ and sample
$l_{i}$ from the following categorical distribution:

\begin{equation}
 l_{i}\sim\mathrm{Categorical}\left(\frac{\left(1-z_{1,n}\right)\exp\left(\left(\mathbf{y}_{n}^{T}\mathbf{w}_{1}\right)^{2}\right)}{\sum_{k}\left(1-z_{k,n}\right)\exp\left(\left(\mathbf{y}_{n}^{T}\mathbf{w}_{1}\right)^{2}\right)},\ldots,\frac{\left(1-z_{K,n}\right)\exp\left(\left(\mathbf{y}_{n}^{T}\mathbf{w}_{K}\right)^{2}\right)}{\sum_{k}\left(1-z_{k,n}\right)\exp\left(\left(\mathbf{y}_{n}^{T}\mathbf{w}_{K}\right)^{2}\right)}\right)
 \label{eq:Parametric_APPCA_Z}
\end{equation}
where after each draw we set $z_{n,l_{i}}=1$.  In dimensionality
reduction applications we often assume $L$ being two or three, hence $l_{1}$
might indicate the $x$-axis, $l_{2}$ the $y$-axis and $l_{3}$
the $z$-axis of the lower dimensional subspace. A Gibbs sampler for
the aPPCA is suggested in Algorithm \ref{alg:Gibbs}.

\begin{algorithm}
\textbf{\small{}Input:}{\small{} $\mathbf{Y},\boldsymbol{\Theta},\mathrm{MaxIter}$}{\small\par}

\textbf{\small{}Initialise:}{\small{} Sample a random $\left(K\times N\right)$
binary matrix $\mathbf{Z}$ and initialize $\mathbf{W}$ using PCA}{\small\par}

\textbf{\small{}for}{\small{} $\mathrm{iter}\leftarrow1$ to $\mathrm{MaxIter}$}{\small\par}

{\small{}$\quad$}\textbf{\small{}for}{\small{} $n\leftarrow1$ to
$N$}{\small\par}

{\small{}$\quad$$\quad$Set $\mathcal{I}=\left\{ k:z_{k,n}=1\right\} $}{\small\par}

{\small{}$\quad$$\quad$}\textbf{\small{}for $l\leftarrow1$ }{\small{}to
$L$}{\small\par}

{\small{}$\quad$$\quad$$\quad$Set $z_{\mathcal{I}_{l},n}=0$}{\small\par}

{\small{}$\quad$$\quad$$\quad$Sample $\mathcal{I}_{l}$ using \eqref{eq:Parametric_APPCA_Z}}{\small\par}

{\small{}$\quad$$\quad$$\quad$Set $z_{\mathcal{I}_{l},n}=1$}{\small\par}

{\small{}$\quad$}\textbf{\small{}for}{\small{} $n\leftarrow1$ to
$N$}{\small\par}

{\small{}$\quad$$\quad$}\textbf{\small{}for}{\small{} $k\leftarrow1$
to $K$}{\small\par}

{\small{}$\quad$$\quad$$\quad$}\textbf{\small{}if }{\small{}$z_{k,n}=1$}{\small\par}

{\small{}$\quad$$\quad$$\quad$$\quad$Sample $x_{k,n}$ using
\eqref{eq:A-PPCAPostX}}{\small\par}

{\small{}$\quad$Sample $\mathbf{W}$ using \eqref{eq:PostW}}{\small\par}

{\small{}$\quad$Sample $\left\{ \sigma^{2},\alpha\right\}$ using Appendix \ref{Appendix:DistributionParameters}}{\small\par}

\caption{Pseudocode for inference in parametric aPPCA using Gibbs sampling.}
\label{alg:Gibbs}
\end{algorithm}

\subsection{Relationship to PCA}
If we marginalize the likelihood (from Equation \eqref{eq:APPCAlikelihood-1}) with respect to the discrete and continuous
latent variables $\left\{ \mathbf{x}_{n},\mathbf{z}_{n}\right\}$)
and take the SVA limit $\sigma^{2}\rightarrow0$,
the maximum likelihood solution
with respect to the transformation
matrix $\mathbf{W}$ is a scaled version of the $K$ largest eigenvectors of the
covariance matrix (like PCA) of the data (multiplied by orthonormal
rotation); proof of this can be seen in Appendix \ref{Appendix:APCA}. Furthermore, different priors over the matrix $\mathbf{Z}$ result in different variants of the model, giving
explicit control over the scale of the different projection axis.

\section{Experiments}

This section provides some empirical results on the performance of the proposed variants of PCA and FA techniques applied to data visualization, data whitening and blind source separation. The methods are evaluated on different kinds of synthetic data, images of handwritten digits from MNIST, images of objects from the Coil-20 dataset, and functional magnetic resonance imaging (fMRI) data.
\begin{table}
\caption{
Performance of factor analysis (FA) methods measured in terms of mean absolute reconstruction error. Different variations of parametric and nonparametric latent feature FA as well as vanilla FA are compared. The FA variations were tested on discrete-continuous synthetic datasets of 1000 points all assuming $\mathbf{Y}=\mathbf{W}\left(\mathbf{X}\odot\mathbf{Z}\right)+\mathbf{E}$,
where $\mathbf{Y}$ is a $\left(D\times1000\right)$ observation matrix,
$\mathbf{X}$ is a $\left(K\times1000\right)$ latent feature matrix, $\mathbf{W}$
is a $\left(D\times K\right)$ factor loading matrix and $\mathbf{E}$
is a $\left(D\times1000\right)$ noise matrix. The latent feature indicator matrix  $\mathbf{Z}$ is $\left(K\times1000\right)$  binary matrix and  $\mathbf{Z}$ is all that changes in the different settings. We have considered 5 separate synthetic sets and the distribution of $\mathbf{Z}$ for each is displayed in Figure \ref{fig:DifferetPriorZComparison}.}

\begin{centering}
\begin{tabular}{|c|c| c|c| c|c| c|c| c|c| c|}
\hline 
{Prior } & \multicolumn{2}{c|}{\textbf{Sparse}} & \multicolumn{2}{c|}{\textbf{Dense}} & \multicolumn{2}{c|}{\textbf{Subspace}} & \multicolumn{2}{c|}{\textbf{Balanced}} & \multicolumn{2}{c|}{\textbf{Single}}\tabularnewline
 & \multicolumn{2}{c|}{\textbf{matrix}}  & \multicolumn{2}{c|}{\textbf{matrix}} & \multicolumn{2}{c|}{\textbf{clustering}} & \multicolumn{2}{c|}{\textbf{matrix}} & \multicolumn{2}{c|}{\textbf{state}}\tabularnewline
\hline 
$K$ & 10 & 20 & 10 & 20 & 10 & 20 & 10 & 20 & 10 & 20\tabularnewline
\hline 
\hline 
Factor analysis &{.012}& {.014} & {.012} & {.014} & {\textbf{.012}} & {.015} & 
{.088} & {.099} & {\textbf{.012}} & {\textbf{.014}}\tabularnewline
 (FA) &  &  &  &  &  &  &  &  &  & \tabularnewline
\hline 
Finite sparse & {.014} & {.018} & {.015} & {.018} & {.016} & {.018} & {.090} & {.101} & {.014} & {.019 }\tabularnewline
 FA &  &  &  &  &  &  &  &  &  & \tabularnewline
\hline 
Infinite sparse & {.023} & {.042} & {.021} & {.042} & {.024} & {.046} & {.073} & {.970}& {.045} & {.075 }\tabularnewline
 FA &  &  &  &  &  &  &  &  &  & \tabularnewline
\hline 
Adaptive FA  & {.014} & {.019} & {.012} & {.019} & {.013} & {.021} & {.047} & {.056}  & {.015} & {.019 }\tabularnewline
(aFA) Gibbs &  &  &  &  &  &  &  &  &  & \tabularnewline
\hline 
Adaptive FA  & {\textbf{.011}} & {\textbf{.013}} & {\textbf{.011}} & {\textbf{.013}} & {\textbf{.012}} & {\textbf{.014}} & {\textbf{.034}} & {\textbf{.044}} & {\textbf{.012}} & {\textbf{.014 }}\tabularnewline
(aFA) EM &  &  &  &  &  &  &  &  &  & \tabularnewline
\hline 
\end{tabular}
\par\end{centering}
\label{tbl:ResultFAcomparison}
\end{table}

\subsection{Synthetic data from latent feature FA models}

\begin{figure}
\includegraphics[scale=0.5]{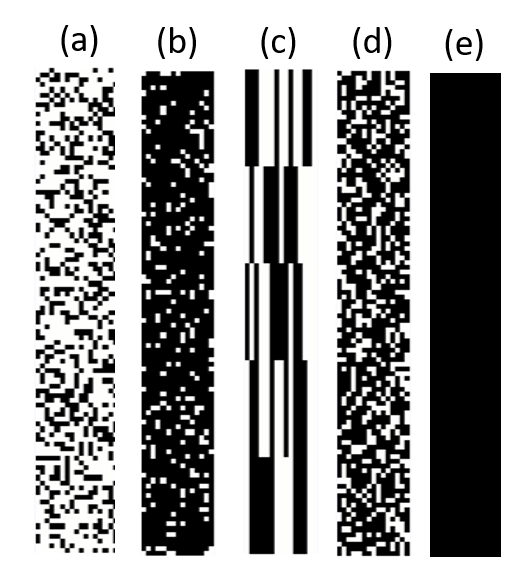}
\centering{}\caption{A plot of the different distributions used to model the latent latent space in Table \ref{tbl:ResultFAcomparison}. The subplots display different samples of the zero-one indicator matrix $\mathbf{Z}$: black cells indicate $1$'s and white cells indicate $0$'s. Five different latent models are considered: (a) Sparse latent feature model, (b) Dense latent feature model, (c) Latent class model in which sharing of some feature between subsets of points implies sharing of all features of those points, (d) Balanced latent feature model sampled from specific hypergeometric distribution, (e) Collapsed latent space consisting of a single state.
 \label{fig:DifferetPriorZComparison}} 
\end{figure}

First, we generate a wide variety of latent feature linear Gaussian datasets, assuming that the data matrix $\mathbf{Y}$ takes the form: $\mathbf{Y}=\mathbf{W}(\mathbf{X}\odot\mathbf{Z})+\mathbf{E}$ with $\mathbf{X}$ a latent feature matrix with standard Gaussian distribution; $\mathbf{W}$
is a factor loading matrix with columns drawn from a multivariate Gaussian with mean zero and covariance matrix $\sigma_{W}^{2}\mathbf{I}_{K}$ with $\sigma_{W} = 1$; $\mathbf{E}$ noise matrix with multivariate Gaussian columns each
with mean zero and covariance matrix $\sigma^{2}\mathbf{I}_{D}$ with $\sigma = 0.1$. The core of the generative model remains the same across the different datasets we generate and only the latent feature indicator matrix  $\mathbf{Z}$ changes. We have considered five separate synthetic sets and the distribution of $\mathbf{Z}$ for each setup is displayed in Figure \ref{fig:DifferetPriorZComparison}. In Table \ref{tbl:ResultFAcomparison}, we evaluate how well four different FA methods (i.e. with changing treatment of $\mathbf{Z}$) perform across each scenario. The resulting FA methods tested are: 

\begin{itemize}
\item Factor analysis (FA): the $\mathbf{Z}$ matrix is full of ones and all factors are shared across all points. 
\item Infinite sparse FA (isFA): the $\mathbf{Z}$ matrix is modelled with an IBP prior
(see Equation \eqref{eq:IBPMatrix}) and most factors are shared only across small overlapping subsets of points.
\item Finite sparse FA (fsFA): the $\mathbf{Z}$ matrix is modelled with a finite Beta-Bernoulli distribution across all points and features.
\item Adaptive FA (aFA): $\mathbf{Z}$ is modelled with a multivariate
hypergeometric prior (see Equation \eqref{eq:Zprior-1-1}).
\end{itemize}
Table \ref{tbl:ResultFAcomparison} also includes a second result for the aFA model when trained using the proposed EM algorithm \ref{alg:MEM}. This was done to distinguish between performance gains due to the model architecture and due to inference method. The results in Table \ref{tbl:ResultFAcomparison} suggest that for sparse latent feature data and for single feature linear Gaussian data, most of the methods perform similarly. The isFA model performs consistently worse than all other methods due to its tendency to overestimate the underlying number of latent features. When we set the concentration parameters of isFA to learn the fixed $K$ number of factors, reconstruction error is higher; if we set concentration parameters so as to infer a umber of factors which is higher than the true generating number of factors $K$, the reconstruction error drops. This effect is similar to the one reported by  \cite{miller2013simple} for Dirichlet process mixtures. 

Vanilla FA performs well in terms of reconstruction error, since it uses all factors to express all points, i.e. vanilla FA learns a lot more loadings then the alternative models with more parsimonious structure. In practice latent feature FA methods are used with larger K then vanilla FA due to the fact that for each factor there is a linear combination of only a small subset of data points. fsFA manages to perform well across most settings, often achieving comparable reconstruction error using a lot sparser factor loadings. However, we see its performance drop substantially for non-sparse balanced latent feature models. Due to the generality of the aFA model, it performs well across all settings, since the latent space structures in the synthetic data are all special cases for the multivariate hypergeometric model. The slightly lower reconstruction error of EM versus Gibbs aFA, suggest convergence to good local optima for the proposed EM scheme and convergence issues of the Gibbs sampler.

\subsection{Factor sharing between MNIST Digits}

\begin{figure}
\includegraphics[scale=0.45]{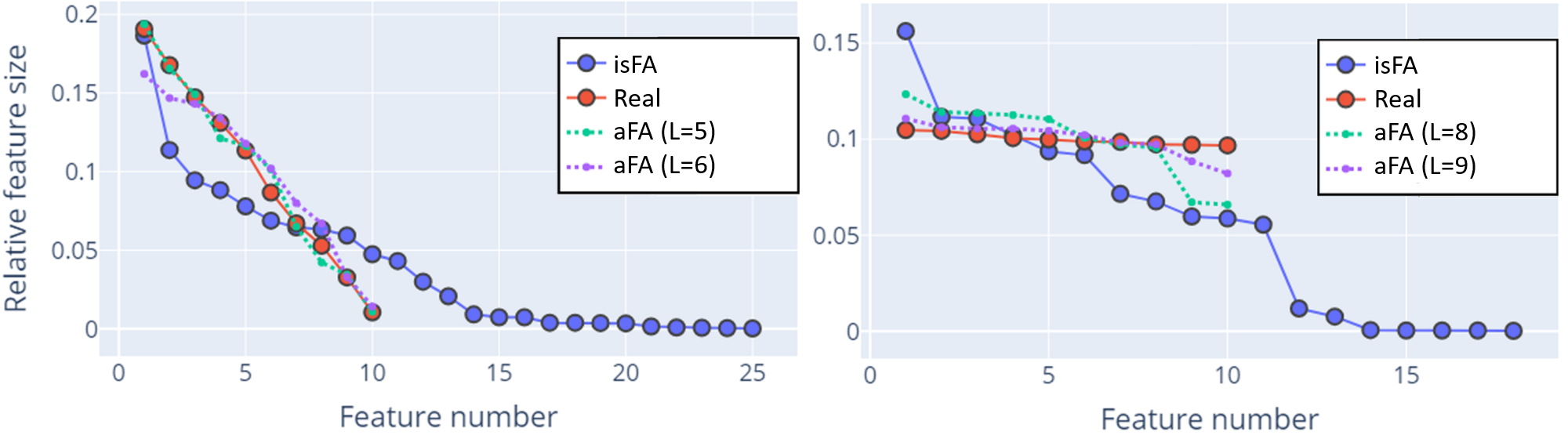}
\caption{ 
Estimated proportion of data associated with the different factors for sparse (left) and dense (right) synthetically generated linear Gaussian data (i.e. as in Table \ref{tbl:ResultFAcomparison}). The $x$-axis denotes the proportion of points associated with a factor (i.e. factor popularity) and the $y$-axis denotes the factor numbers where factors are ordered by size (i.e. number of points associated with them). The true feature popularity is displayed in red; the remaining lines show the feature popularity associated with the estimated factors using the nonparametric factor analysis (isFA) and the proposed adaptive factor analysis  (aFA) model. 
\label{fig:FAPerfomanceComp} }
\end{figure}

In this section we demonstrate training the proposed aFA model on $N = 2500$ odd-labelled digits (500 of each type) from the MNIST handwritten digit dataset. The raw pixel data were first reduced to $D = 350$ using standard PCA since this still preserves $99.5\%$ of the total variance within the data. The total number of unique factors is set to $K=100$ and the number $L$ of observation-specific factors is set to maximize the factor profiles of the different digits. 

\begin{figure}[t]
\includegraphics[scale=0.75]{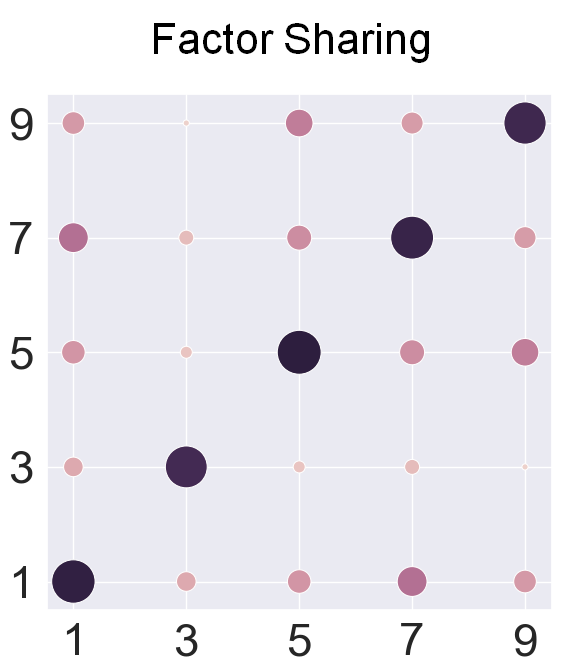}
\includegraphics[scale=0.45]{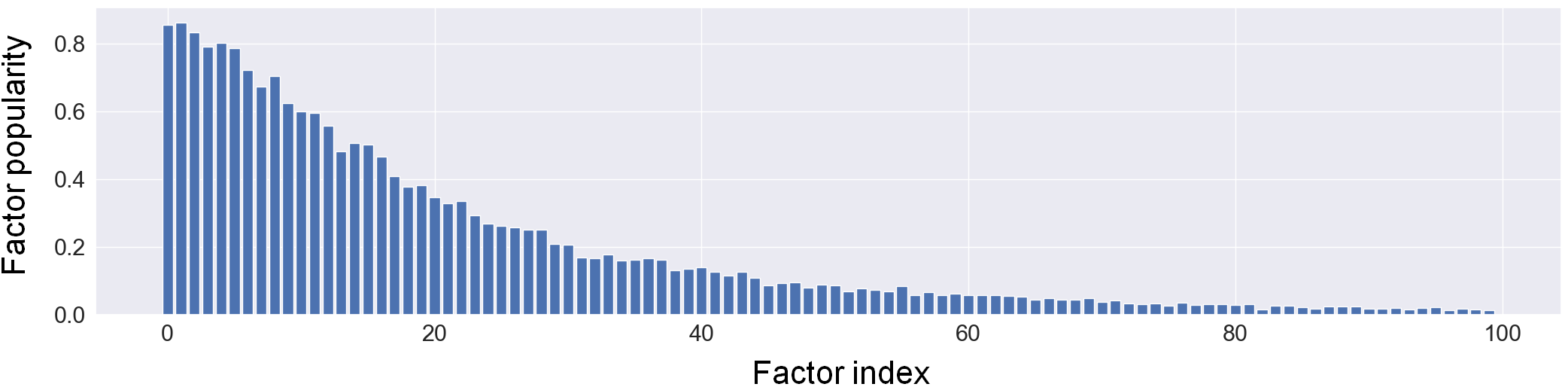}
\centering{}\caption{aFA model trained on 2500 odd-labelled MNIST digits, $500$ of each label. Left: Factor sharing grid between digits: circles are sized depending on the number of features shared between digit pairs denoted on the x-axis and y-axis; color enforces this effect where darker circles indicate more sharing and brighter circles - less. Right: Distribution of feature allocation processes: y-axis denotes the proportion of data sharing the current factor; x-axis indicates the factor number where the factors are ordered based on most popular (left), to least popular with a small number of data points allocated (right).\label{fig:FactorSharing} }
\end{figure}

In Figure \ref{fig:FactorSharing}, we show the factor sharing across the digits which are calculated based on the proportion of factors shared between different digit pairs observations. We count the number of factors shared between samples of 1's and 1's, 1's and 3's, 1's and 5's, 1's and 7's, 1's and 9's, then we normalize by the largest number of features shared; the procedure is repeated for the full grid. The larger and darker circles indicate sharing of more factors. As expected, observations depicting the same digits have the most shared factors; 1's and 7's also share significant structure as well as 5's and 9's which broadly coincides with the geometry of the digits. The results can be directly compared with a similar experiment in \cite{paisley2009nonparametric}. In Figure \ref{fig:FactorSharing} we display the estimated feature weights obtained by summing over the $Z$ matrix and normalizing. Varying $L$ and $K$ one can study how well sparse and dense aFA models infer features specific to the different digits.

\subsection{Visualization with aPPCA}
Despite the increased popularity of nonlinear manifold embedding algorithms for data visualization, linear dimensionality reduction methods remain of fundamental importance to exploratory data visualization, arguably due their scalability, stability, and intuitive data representation. In this section, we provide simple illustrations of how latent feature PCA complements conventional PCA visualizations. 

\begin{figure}[t]
\includegraphics[scale=0.75]{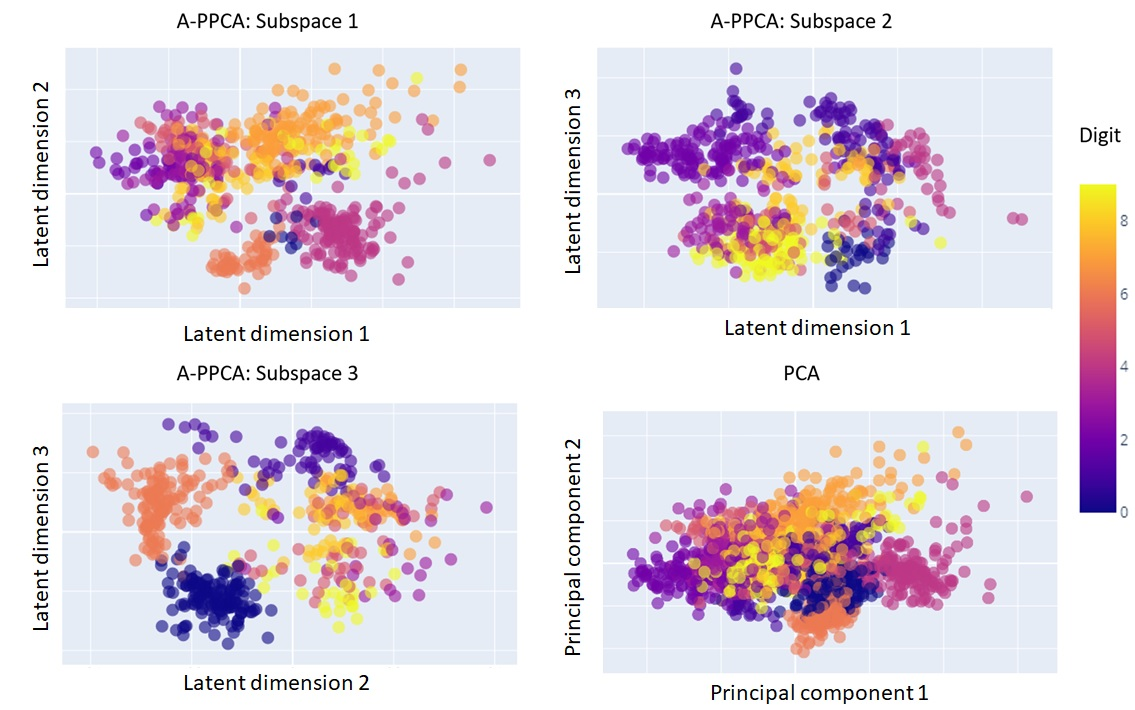}
\centering{}\caption{ Scatter plot of the $2$-dimensional projections of $10,000$ MNIST digits, obtained using aPPCA and PCA. The first $3$ subplots contain only proportions of the data which have been estimated by aPPCA to lie in the corresponding subspace (i.e. Subspace $1$ is spanned by features $1$ and $2$; Subspace $2$ by features $2$ and $3$; Subspace $3$ by $1$ and $3$). The $4$-th subplot shows the $2$-dimensional projection of all digits obtained using PCA. 
 \label{fig:hand-digit}}
\end{figure}

Typically we use PCA to project all of the data down to the first 2 PCs, in aPPCA each point is also reduced to say $L=2$ components, but these components can be computed only based on some of the data, having some larger $K$ unique sparse PCs in total. This essentially means we visualize the data using multiple scatter plots including different subsets of the projected data, instead of the single crowded plot in PCA.

\begin{figure}
\includegraphics[scale=0.45]{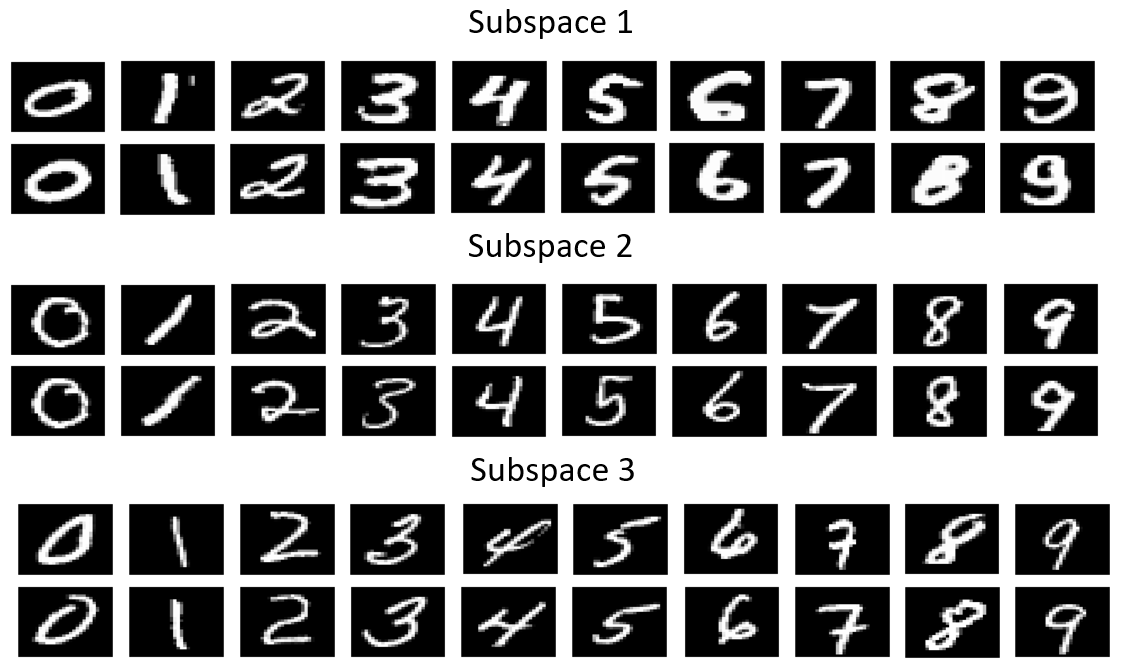}
\centering{}\caption{Randomly selected MNIST digits from each of the identified subspaces. The top panel consist of mostly thicker  digits; the bottom panel is dominated by slanted digits.  \label{fig:behavior_of_MNIST} }
\end{figure}

\paragraph{MNIST Dataset}
First, we look at subspace sharing of MNIST digits. Note that with PPCA we project each data point down onto the same two orthogonal PCs preserving most variance and we display the projections in a single 2-dimensional plot. With aPPCA we can still project each point onto $L = 2$ orthogonal PCs, but the components are not all constrained to be shared for all of the data if $K>L$. For more intuitive visualization, we first use a 2-layer multilayer perceptron variational autoencoder (VAE) \cite{kingma2013auto} to reduce the dimension of 10,000 MNIST digits. The 784-dimensional data is reduced with the VAE to 10 dimensions and then we train parametric aPPCA with $K=3$ and $L=2$ to visualize the digits in the latent space. We will assume that subspace $1$ is spanned by the inferred features $1$ and $2$; subspace $2$ by features $2$ and $3$; subspace $3$ by features $1$ and $3$. Note that all pairs of subspaces share one of their principal axes. In Figure \ref{fig:hand-digit} we display the reduced data in each of these subspaces where we can see increased separation between many of the distinct clusters of different digits. From Figure \ref{fig:behavior_of_MNIST} we can see that distinct geometric properties of digits are encoded in the identified subspaces. Figure \ref{fig:behavior_of_MNIST} shows randomly selected digits from each subspace and we can see that most digits in subspace $1$ are written in thicker font; most digits in subspace $3$ are slanted.  

The visualization reduces the crowding effect of PCA and produces multiple two-dimensional plots which jointly decompose the data and intuitively organize the observed data.

\paragraph{COIL-20 Dataset}
We consider another data visualization example, this time using data from the Columbia University Image Library (COIL-20) \cite{nene1996columbia}. The dataset contains low resolution images ($32\times32$ pixels) of 20 different objects. The objects are placed on a motorized turntable against a blank background and the turntable is rotated through 360 degrees to vary object pose with respect to a fixed camera. 72 images of each object are taken, at pose intervals of 5 degrees rotation and the images are size normalized. This means that objects which are very similar at different view angles, will result in very similar 72-image observations.  

\begin{figure}
\begin{subfigure}{.32\textwidth}
  \centering
  \includegraphics[width=.95\linewidth]{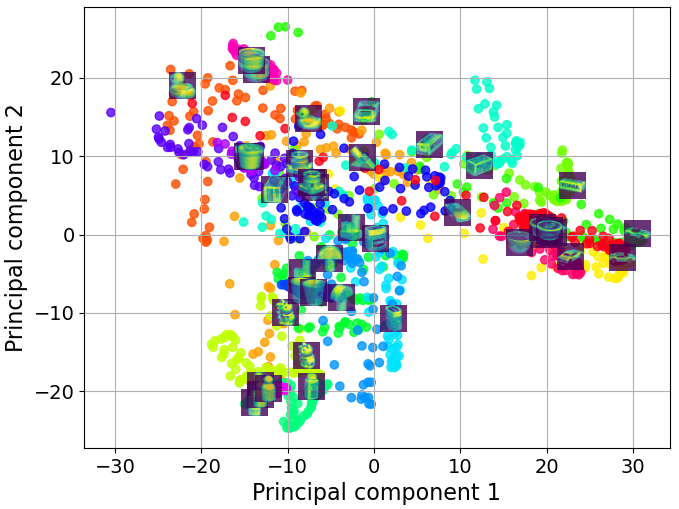}
  \caption{PCA}
  \label{fig:COIL20a}
\end{subfigure}%
\begin{subfigure}{.32\textwidth}
  \centering
  \includegraphics[width=.95\linewidth]{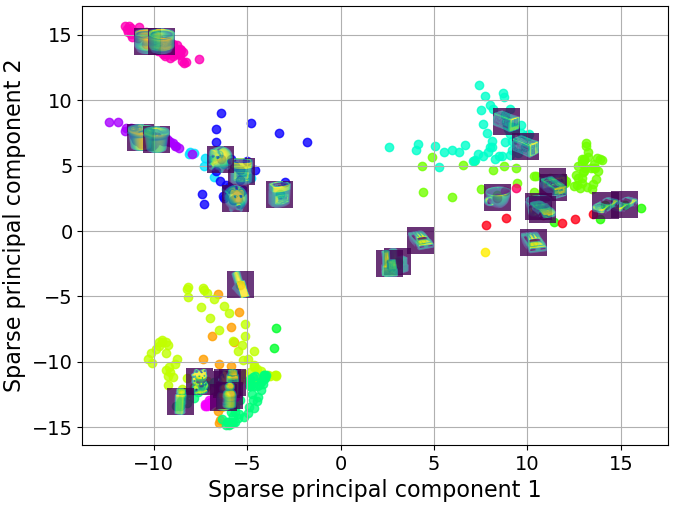}
  \caption{aPPCA: subspace $\{1,2\}$}
  \label{fig:COIL20b}
\end{subfigure}
\begin{subfigure}{.32\textwidth}
  \centering
  \includegraphics[width=.95\linewidth]{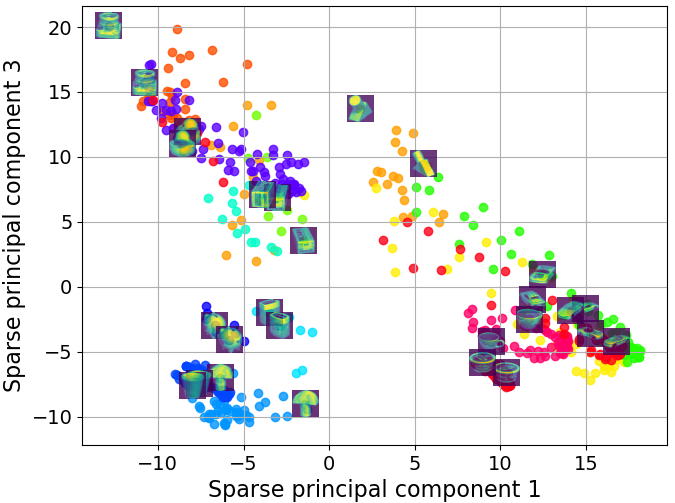}
  \caption{aPPCA: subspace $\{1,3\}$}
  \label{fig:COIL20c}
\end{subfigure}
\caption{2-D projections of the COIL-20 dataset images using PCA and aPPCA methods. (a) shows the 2-D scatter plot obtained by reducing the 1024-D images to 2-D with PCA, a sample of the original images is placed over their projection. (b)-(c) show the 2-D projections of data points onto sparse principal components they are associated with, inferred using aPPCA. Where (a) includes all data points in a single projection, aPPCA in (b)-(c) identifies subsets of the data sets sharing principal components, hence principal components are estimated using only a subset of the observations (i.e. sparse principal components).}
\label{fig:COLI20}
\end{figure}

First, we reduce all the 1440 images onto the two PCs which are computed to preserve the variance globally across all data points. Images from the different objects are displayed in different colors in Figure \ref{fig:COLI20}, whereas a fraction of the actual images is overlaid on the scatter plot. We see that some of the objects, such as two of the toy cars framed from front view angle (i.e. green and yellow class on the far right of the plot), are well separated with other rectangular objects with similar geometry. However, most of the objects are bundled in the center of the plot and not recognizable in the reduced 2-dimensional space.

Next, we fit an aPPCA model with $K=4$ and $L=2$ which effectively learns  four sparse PCs with each data point associated a subset of exactly two of these components. In (b)-(c) of Figure \ref{fig:COLI20}, we display the points sharing combinations of the estimated subspaces spanned by the sparse principal components (i.e. four unique  components leads to $(4\times3)/2=6$ subspaces with some shared axes). We see that projections onto the sparse PCs reduce the crowding effect of PCA. In addition, the different sparse PCs encode interpretable geometric properties of the objects observed. For example, objects with smaller values along the sparse PC number two, tend to be more narrow, whereas, objects with large values along the sparse principal component 1 tend to be less cylindrical. Within each 2-D subspace, the different object projections are easier to separate and different objects with similar projections also have intuitive image similarity under a rotation angle. 

\paragraph{Interpreting global structure in manifold embedding} Toy problems such as COIL-20 have been used to showcase manifold embedding methods such as t-SNE \cite{Maaten2008} and more recently UMAP \cite{mcinnes2018umap}. Empirically, both t-SNE and UMAP often lead to very good class separability in the lower dimensional projections particularly in scenarios when class separability in the original high-dimensional data is good (i.e. such as for COIL-20). At the same time, it is well known that many manifold embedding algorithms such as UMAP and t-SNE do not preserve the global structure of the data manifold, unlike linear methods such as PCA and multidimensional scaling, or kernel space models like the Gaussian process latent variable models.

\begin{figure}
\begin{subfigure}{.50\textwidth}
  \centering
  \includegraphics[width=.95\linewidth]{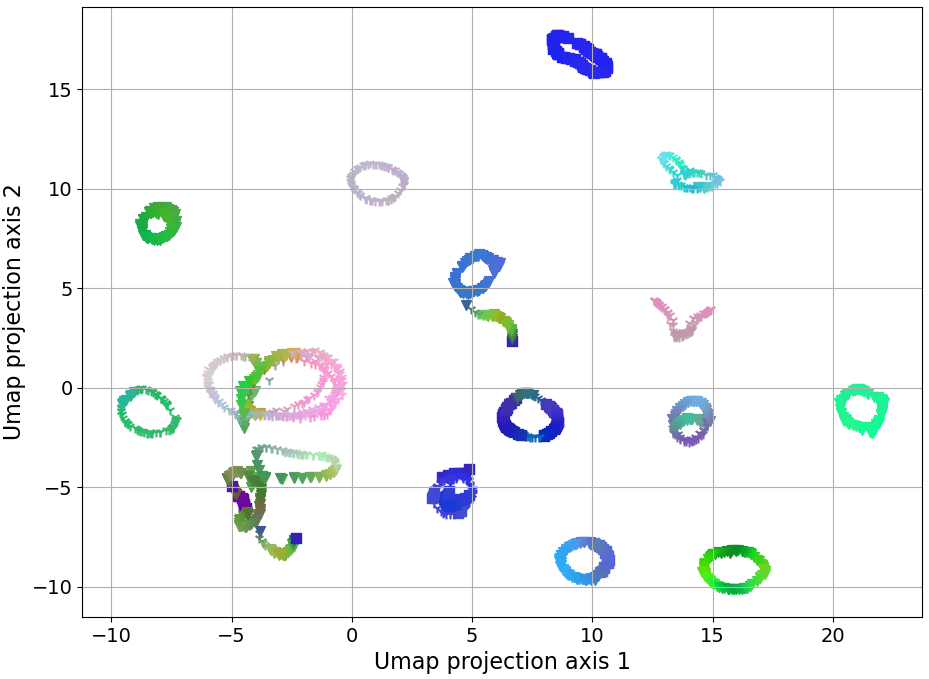}
  \caption{UMAP: colored by aPPCA}
  \label{fig:sfig1_umap}
\end{subfigure}%
\begin{subfigure}{.50\textwidth}
  \centering
  \includegraphics[width=.95\linewidth]{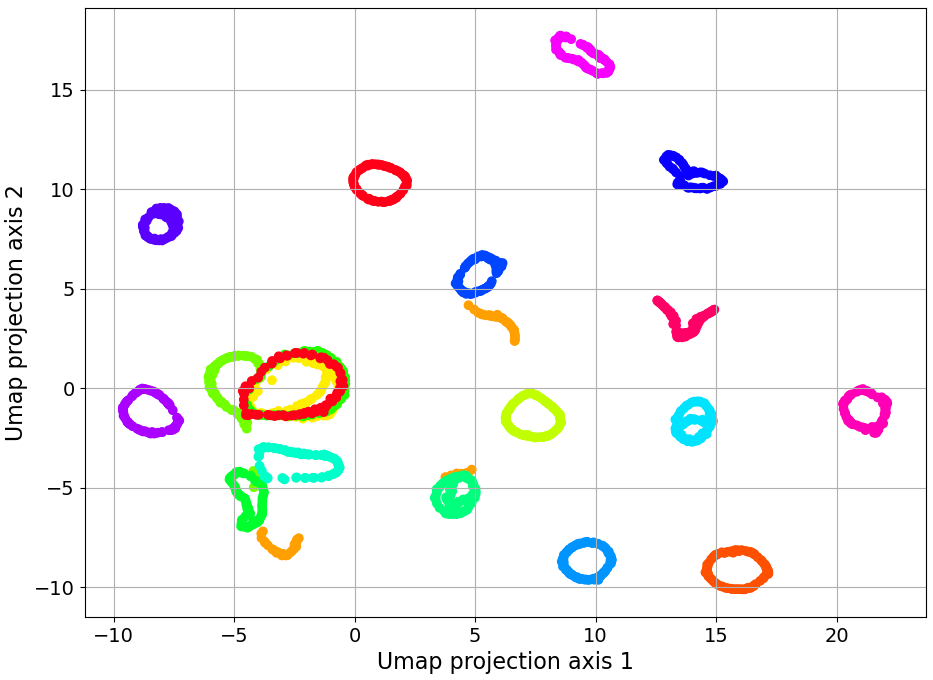}
  \caption{UMAP: colored by classes}
  \label{fig:COLI20_diagnostic_sub}
\end{subfigure}
\caption{2-D projections of the COIL-20 dataset images using UMAP. The $x$-axis and $y$-axis are determined based on the UMAP projection. In (a) the colors encode the object class of each point. In (b) the colors encode the 3-D projection of the points done via aPPCA. Each point is associated with exactly three sparse PCs, but the total number of components is larger. Points that also share the same subspace (i.e. they are associated with the same three sparse PCs) are plotted with the same symbol. Note that no two points have exactly the same color and color similarity indicates only proximity.}
\label{fig:COLI20_diagnostic}
\end{figure}

This often leads to lower dimensional projections which reflect well class separability when captured in localized regions of the manifold (like in MNIST and COIL-20), but do not capture similarities across different classes adequately. To illustrate, Figure \ref{fig:COLI20_diagnostic} shows 2-D projections of COIL-20, obtained using UMAP. On the right (in Figure \ref{fig:COLI20_diagnostic_sub}), are objects associated with each class which have been also color coded. Most objects are well separated into distinct clusters with little overlap across objects, except for the different car images and the package images in the center of the figure. Certain objects have been separated into 2 or 3 clusters (i.e. the duck and the bowl), depending on the angle of view, but if the aim is object classification based on 2-D embedding of the data, the task is nearly trivial. The challenge is less clear if we are looking to uncover latent structure between the objects. 

\begin{figure}
    \centering
    \includegraphics[scale=0.70]{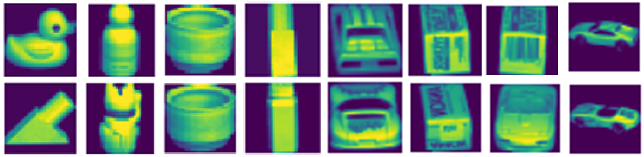}
    \caption{Example images from different object classes in the COIL-20 dataset, with shared subspaces and proximity in the 3-D orthogonal aPPCA projection of input images. Proximity was defined with basic K-means clustering of the lower dimensional projections, where Figure \ref{fig:COLI20} shows how clustered specific subspaces are. Note that objects sharing subspaces are merely estimated to shared covariance structure.  }
    \label{fig:three graphs}
\end{figure}

\cite{mcinnes2018umap} has suggested using PCA to reduce data onto its first three PCs and color UMAP embeddings using RGB values defined by the 3-D PCA projections of each point. This approach suggests that points close in the PCA projection of the data, would also have a similar color. By contrast, as colors transition, this means that data points are projected far apart on some of the PCs. The problem with using PCA as diagnostics for UMAP projections in this manner, is that we are likely to crowd observations overestimating proximity between most points due to the simplistic assumptions of PCA. If we are interested in using manifold embedding methods such as UMAP which preserve the local structure of the original manifold, we could use piecewise linear methods such as aPPCA which capture the global structure of the manifold and use these to annotate the 2-D UMAP projections as seen in Figure \ref{fig:COLI20_diagnostic_sub}. In Figure \ref{fig:COLI20_diagnostic_sub}, we use different symbols to denote points associated with different subspaces; the colors depend on the 3-D projection obtain with a single run of aPPCA with $K=4$ and $L=3$ (i.e. leading to four subspaces spanned by sparse PCs $\{1,2,3\}$; $\{2,3,4\}$; $\{1,2,4\}$ and $\{1,3,4\}$). Note that under this diagnostic, similar colors (in RGB values) indicate similarity in the reduced form. We can see that aPPCA much of the omitted cross-object similarities specific to certain rotations such as between: the rotated Maneki-neko (i.e. lucky cat figurine) and cylindrical bottle; the duck toy and the similar shape wooden part; the different clusters of bowl images and others. For more intuition we have also included images of example rotated object similarities identified using subspace decomposition diagnostics with aPPCA, see Figure \ref{fig:three graphs}.

\subsection{Data pre-processing}
Another ubiquitous use of PCA is \textit{data whitening}. This is an often used pre-processing step which aims to \textit{decorrelate} the observed data to simplify subsequent processing and analysis, for example, image data tends to have highly correlated adjacent pixels. In this capacity, PCA works by ``rotating'' the data in observation space, retaining dimensionality unlike with visualization applications.

Here we show a simple example demonstrating how aPPCA can be used to do more effective \textit{local} whitening which can lead to more accurate and interpretable supervised classification in decorrelated latent feature space. To demonstrate this, we compare a classifier trained on raw data with the same classifer trained on the first few PC projections of the data where the PCs are estimated (1) globally using PCA and (2) locally, within subsets of the data using aPPCA.

\begin{figure}
\includegraphics[scale=0.65]{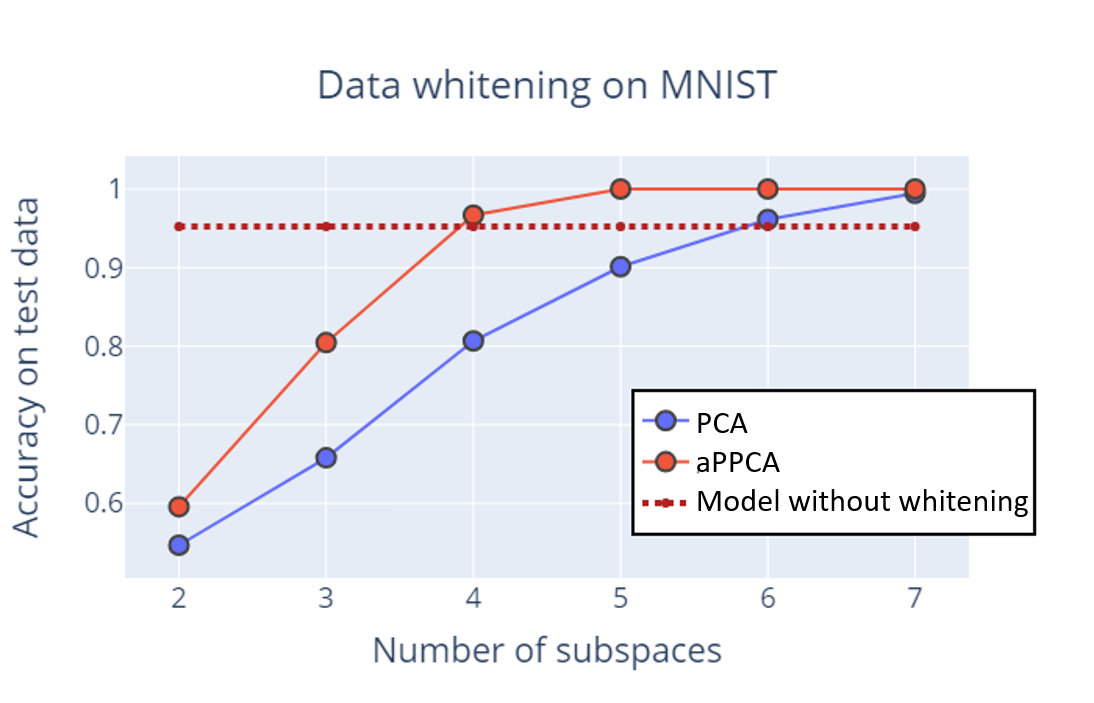}
\centering{}\caption{\label{fig:whiten}
Classification accuracy of a multilayer perceptron evaluated using 10,000 MNIST digits in three different setups: no whitening; data pre-processed with PCA; data pre-processed with adaptive probabilistic PCA (aPPCA). On the $x$-axis we show the number of reduced dimensions for different instances of the same classifier. The $y$-axis indicates the out-of-sample accuracy, evaluated using $10$-fold cross-validation.  }
\end{figure}

For simplicity, we show an example of pre-processing the MNIST handwritten
digit classification dataset, before training a multilayer perceptron. We train a simple
multilayer perceptron with one hidden layer with a softmax activation function on a $9000$-image subset
of the $784$-D MNIST dataset with 1000 images reserved for testing.
We compare the performance of the same classifier network when (1)
trained on the original $784$-D pre-processed data, (2) trained on lower dimensional projection of the data using PCA (3) trained on
  data locally whitened by aPPCA ($K$-dimensional). The classifier is a multilayer perceptron in all three scenarios. Figure \ref{fig:whiten} shows the classification
accuracy of these three different pre-processing approaches as we
vary $K$, i.e. the number of PCs onto which we can project the data down. For aPPCA, we have kept $L=K-1$ for simplicity. Intuitively, we also see
increases in performance if multiple, separate classifiers are trained
on each $L$-dimensional subspace, but usually, after whitening with PCA, a single classifier is used.

A key feature of the aPPCA algorithm for localized data whitening is that it estimates more robust subspaces which can be seen in the smaller number of subspaces (i.e. PCs or columns of $\mathbf{W}$) required for training of the same classifier, to achieve better out-of-sample performance. The multilayer perceptron trained on PCA whitened data requires more subspaces in training to achieve comparable out-of-sample performance.

\subsection{Blind source separation in fMRI}

Functional magnetic resonance imaging (fMRI) is a technique for the non-invasive study of brain function. fMRI can act as an indirect measure of neuronal activation in the brain, by detecting blood oxygenation level dependent (BOLD) contrast \cite{Zarahn1997}. BOLD relies on the fact that oxygenated (diamagnetic) and deoxygenated (paramagnetic) blood have different magnetic properties. When neurons fire there is a resultant increase in localised flow of more oxygenated blood, which can be detected using BOLD fMRI.

\begin{figure}[t]
\includegraphics[scale=0.80]{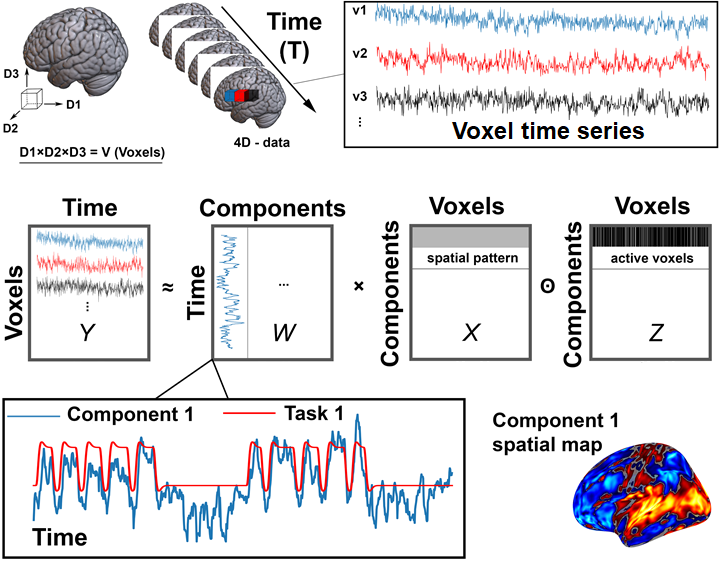}
\centering{}\caption{\label{fig:Time_course}
fMRI data of 3-D brain volumes collected over time (e.g. every 1 second). Typically, images are vectorised and represented as 2-D $\mathbf{T}\times \mathbf{V}$ matrices (top panel), with $\mathbf{V}$ being number of all voxels in all dimensions and $\mathbf{T}$ the number of time instances. This matrix can be then reduced down to a $\mathbf{K}\times \mathbf{V}$ matrix (i.e. $\mathbf{X}$) which represents spatial maps of regions with intrinsically similar time-courses (middle panel). $\mathbf{W}$ denotes the modelled transformation matrix and $\mathbf{Z}$ indicates whether components (i.e. rows of $\mathbf{X}$) should be included in the representation of the data matrix or not. Columns of $\mathbf{W}$, also referred to as components, are easier to interpret in terms of their correlation to experimental stimuli.}
\end{figure}

fMRI time-series data is often represented as a series of three-dimensional images (see Figure \ref{fig:Time_course}). However, data can be also represented as a two-dimensional matrix using vectorized voxel matrices over time (time by voxels). In this representation each matrix row contains all voxels from the brain image (or the subset selected for analysis) from a single time instance. Although useful, fMRI data often suffers from low image contrast-to-noise ratio, it is biased by subject head motions, scanner drift (i.e. due to equipment overheating) and from signals from irrelevant physiological sources (cardiac or pulmonary). Therefore, direct analysis of raw fMRI measurements is rare \cite{Pruim2015} and domain experts tend to work with pre-processed, reduced statistics of the data. In clinical studies, due to the typical scarcity of fMRI series per subject and the low signal-to-noise ratio, flexible black-box algorithms are rarely used. The preferred methods for pre-processing of fMRI series and localization of active spatial regions of the brain are variants of linear dimensionality reduction methods such as PCA and FA  \cite{Calhoun2003,Taghia2017bayesian,Beckmann2005,Pruim2015,hojen2002analysis}. Typically, of primary interest is then analysis of a representative subset of the inferred PCs or factors respectively, instead of the use of raw data.

A key problem with this approach is that these linear methods assume that the components/factors are a linear combination of all of the data, i.e. in other words, PCA and FA assume that all components are \textit{active} for the full duration of the recording. Common implementations for fMRI series \cite{mckeown2003independent,calhoun2009review} might adopt thresholding the inferred components or using sparse versions of the decomposition techniques. These can still lead to biased decomposition into components and we are likely to overestimate the firing area of the brain for some components and completely overlook functional areas of the brain which are active for short periods of time. Here, we show that our proposed \textit{adaptive} linear methods, are better motivated models for alleviating this problem and can infer better localized spatial regions of activation from fMRI. Furthermore, we can potentially discover novel short-term components in a principled, probabilistic, data-driven fashion. 

\begin{figure}[t]
\includegraphics[scale=0.50]{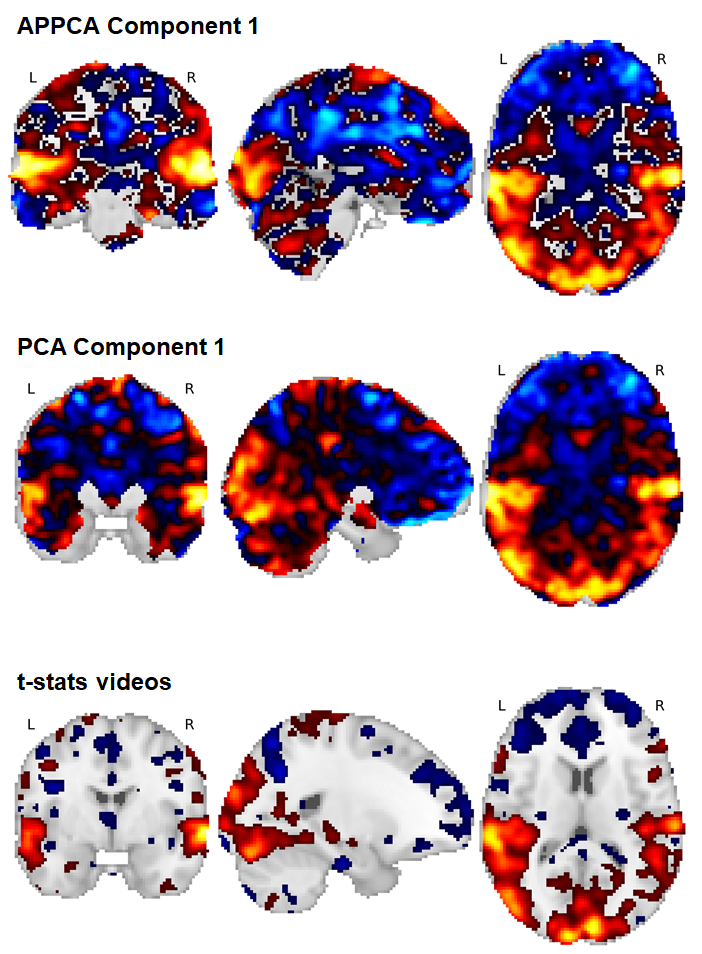}
\centering{}\caption{\label{fig:fMRI} Lower dimensional fMRI recording reduced across time, plotted against the subject brain. The fMRI time series of length $T$ is reduced to $K$ components and here we display the single component most associated with the stimuli during the experiment. The top panel displays the reduced projection estimated using aPPCA and the middle panel the projection estimated using PPCA. The larger amount of grey regions indicates that aPPCA projection better localizes the regions of the brain fluctuation through time, as a response to the visual stimuli. Reference regions of activation can be seen from the \textit{t-map} in the bottom panel displaying the correlation of the component with the ground-truth visual stimuli.}
\end{figure}

As a proof of concept, here we apply aPPCA to fMRI data collected from a single participant while exposed to continuous visual stimuli. fMRI data was initially realigned to correct for subject motion and registered to a group template (Montreal Neurological Institute Template). Using a 3T Siemens scanner, a whole brain image with voxel resolution of  $2\times2\times2$ mm was acquired each 0.8 seconds. The data had $215,302$ voxels and $989$ time instances. aPPCA decomposition was performed by treating time instances as features, which is a standard procedure in the neuroimaging field. For aPPCA we used $K = 500$ unique components and constraint of $L = 200$ components, which were selected to achieve component similarity with the benchmark and enable visually intuitive comparisons. We also performed PPCA with $K = 200$  components for comparison, see  Figure \ref{fig:fMRI}.

The figure shows the component most associated with the task estimated both with aPPCA and PPCA. aPPCA results in sparser maps across space, which enhance localization. This sparsity increases with higher numbers of components that explain less variance in the data. This can be useful for identifying noisy components and brain areas that are only transiently active during task performance. We also show the corrected t-statistic map (Figure \ref{fig:fMRI}) which shows the voxels that have significant correlation with the visual stimuli. The map is family-wise error (FWE) rate corrected at $p<0.05$ at voxel threshold $p<0.001$. One benefit of decomposition methods versus standard correlation methods is that they do not need a predefined model of assumed task activation.

Direct quantitative evaluation of pre-processing tools for fMRI data is an open problem, due to the lack of clear ground-truth definition of brain-activity related components. We have measured the mean reconstruction error across all $215,302$ voxels as well as the standard deviation across voxels. We find that highest error with highest standard deviation (i.e. average root mean square error (RMSE) of \textbf{16.5}, standard deviation of RMSE of \textbf{4.8}) was obtained using PPCA. aPPCA reconstruction gradually reduces these errors depending upon the ratio of $K$ and $L$ used, with the best scoring reconstruction having average RMSE of \textbf{14.1} and standard deviation RMSE (across voxels) of \textbf{3.0}. The lower standard deviation of error across voxels supports our hypothesis of better preserved local region information using aPPCA. Due to the simplicity of the imaging setup, both methods were able to identify components highly correlated to the stimuli, see Figure \ref{fig:fMRI}. The typical goal for experts would be to examine functions of the specific brain regions or networks, as well as, potentially affected areas of the brain after head trauma or stroke. 

The common analysis practice would be to threshold the observation specific loadings (i.e. reduced form data) and only consider voxels that \textit{significantly} contribute to selected subsets of components. The adaptive nature of aPPCA allows us to infer the voxels association with specific components (i.e. $\mathbf{Z}$ switches off voxels not part of a component) in a principled fashion as a part of a fully probabilistic model. In addition, the experimental user has explicit control over the contrast voxels used in different components (ratio of $K$ and $L$) and this can be useful for achieving better spatial localization, without thresholding which is an inherently subjective procedure.

\section{Summary and conclusions}

In this work, we have studied generic discrete latent variable augmentation for ubiquitous linear Gaussian methods applied for feature learning, whitening and dimensionality reduction applications. The manuscript details some shortcomings with existing Bayesian nonparametric linear Gaussian methods and demonstrates that flexible alternatives can be derived using latent hypergeometric distributions. This leads to our novel aFA and aPPCA models which be trained efficiently, yet overcome the inherent over-partitioning in Beta processes and allows for more flexible regularization of the model capacity, compared to Beta-Bernoulli models. The proposed models can be extended to many other related methods such as generalized linear Gaussian models, Gaussian process latent variable models (GPLVMs), kernel PCA methods, and others. \cite{dai2015spike} has already introduced the problem of handling discontinuity in GPLVMs and proposed a simple \textit{spike and slab prior} to augment the continuous latent variables in GPLVMs. Augmenting GPLVMs with discrete hypergeometric feature allocation indicators, would in principle, allow for a richer and more compact model of the manifold using a smaller number of underlying, feature-specific Gaussian processes. In our study of aPPCA models, we have also proposed efficient practical inference methods for distributions on Stiefel manifolds.
The utility of the proposed tools is demonstrated on a wide range of synthetic latent feature Gaussian data sets, MNIST handwritten digit images, COIL-20 object images and brain imaging fMRI data. The synthetic data study shows that a wide range of feature allocation distributions can be captured with a multivariate hypergeometric model. We have applied aPPCA to MNIST variational autoencoder projections, to show that it can be used to identify images sharing clear geometric features. aFA was applied to nearly raw digits to show that images of visually similar digits share more factors than visually distinct digits. We conclude with an application of aPPCA to a widely-encountered problem in brain imaging with fMRI, and demonstrate accurate decomposition of active spatial regions in the brain during different stimuli (or at rest). We also demonstrate that this discrete-continuous decomposition leads to more accurate localization of active brain regions. This finding has the potential to lead to significant improvements to analysis pipelines for fMRI data for neurological screening and cognitive neuroscience applications.

\newpage

\bibliographystyle{plain}
\bibliography{references.bib}

\newpage
\appendix 

\section{Adaptive PCA}\label{Appendix:APCA}

In this section we demonstrate that the proposed aPPCA
model from Section \ref{sec:The-adaptive-probabilistic PCA} is indeed
a generalization of the ubiquitous PCA and using \textit{small variance asymptotics} \cite{broderick2013mad}. Let us first start by marginalizing out the discrete and continuous
latent variables $\left\{ \mathbf{x}_{n},\mathbf{z}_{n}\right\} $
which are not of explicit interest in conventional PCA approach. To
compute the marginal likelihood of $\mathbf{y}_{n}$ we compute the
expectations:
\begin{equation}
\mathbb{E}_{\mathrm{P}\left(\mathbf{x}_{n},\mathbf{z}_{n}\right)}\left[\mathbf{y}_{n}\right]\quad\mathrm{and}\quad\mathbb{E}_{\mathrm{P}\left(\mathbf{x}_{n},\mathbf{z}_{n}\right)}\left[\left(\mathbf{y}_{n}-\mathbb{E}\left[\mathbf{y}_{n}\right]\right)\left(\mathbf{y}_{n}-\mathbb{E}\left[\mathbf{y}_{n}\right]\right)^{T}\right]
\end{equation}
where we use $\mathbb{E}\left[\ \right]=\mathbb{E}_{\mathrm{P}\left(\mathbf{x}_{n},\mathbf{z}_{n}\right)}\left[\ \right]$
for notational convenience. We express the moments of the marginal
likelihood starting with the posterior mean of the marginal, $\mathbb{E}\left[\mathbf{y}_{n}\right]$:
\begin{align*}
\mathbb{E}\left[\mathbf{y}_{n}\right] & =\mathbb{E}\left[\mathbf{W}\left(\mathbf{x}_{n}\odot\mathbf{z}_{n}\right)+\boldsymbol{\mu}+\boldsymbol{\epsilon}_{n}\right]\\
 & =\mathbf{W}\left(\mathbb{E}\left[\mathbf{x}_{n}\right]\odot\mathbb{E}\left[\mathbf{z}_{n}\right]\right)+\boldsymbol{\mu}+\mathbb{E}\left[\boldsymbol{\epsilon}_{n}\right]\\
 & =\mathbf{W}\left(0\odot\boldsymbol{\rho}\right)+\boldsymbol{\mu}+0\\
 & =\boldsymbol{\mu}
\end{align*}
where we have used a diagonal $\left(K\times K\right)$ matrix $\boldsymbol{\rho}$
to denote the expectation of each feature, which is determined by
the prior on the matrix $\mathbf{Z}$:
\begin{equation}
\rho_{k,k}=\begin{cases}
\frac{L}{K} & \mathrm{if\ multivariate\ hypergeometric\ prior}\\
\frac{1}{N}\sum_{n}z_{k,n} & \mathrm{if\ IBP\ prior}
\end{cases}    
\end{equation}
For the variance of the marginal, we can write:
\begin{align*}
\mathbb{E}\left[\left(\mathbf{y}_{n}-\mathbb{E}\left[\mathbf{y}_{n}\right]\right)\left(\mathbf{y}_{n}-\mathbb{E}\left[\mathbf{y}_{n}\right]\right)^{T}\right] & =\mathbb{E}\left[\left(\mathbf{W}\left(\mathbf{x}_{n}\odot\mathbf{z}_{n}\right)+\boldsymbol{\epsilon}_{n}\right)\left(\mathbf{W}\left(\mathbf{x}_{n}\odot\mathbf{z}_{n}\right)+\boldsymbol{\epsilon}_{n}\right)^{T}\right]\\
 & =\mathbf{W}\boldsymbol{\rho}\mathbf{W}^{T}+\sigma^{2}\mathbf{I}_{D}
\end{align*}    
Finally, using the obtained expression for $\mathbb{E}\left[\mathbf{y}_{n}\right]$
and $\mathbb{E}\left[\left(\mathbf{y}_{n}-\mathbb{E}\left[\mathbf{y}_{n}\right]\right)\left(\mathbf{y}_{n}-\mathbb{E}\left[\mathbf{y}_{n}\right]\right)^{T}\right]$,
combined with the Gaussian likelihood of $\mathbf{y}_{n}$ resulting
in a linear Gaussian model, we can write the marginal likelihood as:
\begin{equation}
\mathrm{P}\left(\mathbf{y}_{n}\mid\mathbf{W},\boldsymbol{\rho},\sigma\right)=\frac{1}{\left(2\pi\right)^{\frac{D}{2}}}\left|\mathbf{C}\right|^{-1/2}\exp\left(-\frac{1}{2}\mathbf{y}_{n}^{T}\mathbf{C}^{-1}\mathbf{y}_{n}\right)
\label{eq:Margianal}
\end{equation}
where we used $\mathbf{C}=\mathbf{W}\boldsymbol{\rho}\mathbf{W}^{T}+\sigma^{2}\mathbf{I}_{D}$
to denote the model covariance. 

Now, the marginal likelihood in this collapsed aPPCA model is almost
identical to the PPCA model Tipping and Bishop (1999b) with the key
difference being the weights $\boldsymbol{\rho}$ which can be scalar
shared across each dimension or direction specific. In fact, we can
say that the PPCA model is a special case of the collapsed aPPCA model
when the diagonal of $\boldsymbol{\rho}$ are full of ones, which
occurs when the matrix $\mathbf{Z}$ is full of ones implying all
observations are active in all $K$ number of one-dimensional subspaces. 

The complete data log-likelihood of the collapsed model is:
\begin{align*}
\mathcal{L} & =\sum_{n=1}^{N}\ln\left(\mathrm{P}\left(\mathbf{y}_{n}\mid\mathbf{W},\boldsymbol{\rho},\sigma\right)\right)\\
 & =-\frac{N}{2}\left(D\ln\left(2\pi\right)+\ln\left|\mathbf{C}\right|+tr\left(\mathbf{C}^{-1}\mathbf{S}\right)\right)
\end{align*}    
where $\mathbf{S}=\frac{1}{N}\mathbf{Y}\mathbf{Y}^{T}$. To find the
maximum likelihood estimate for $\mathbf{W}$, we differentiate the
likelihood and solve:
\begin{equation}
\frac{d\mathcal{L}}{d\mathbf{W}}=-\frac{N}{2}\left(2\mathbf{C}^{-1}\mathbf{W}\boldsymbol{\rho}-2\mathbf{C}^{-1}\mathbf{S}\mathbf{C}^{-1}\mathbf{W}\boldsymbol{\rho}\right)=0
\end{equation}
The maximum likelihood estimate for $\mathbf{W}$ then should satisfy:
\begin{align*}
\mathbf{C}^{-1}\mathbf{W}\boldsymbol{\rho} & =\mathbf{C}^{-1}\mathbf{S}\mathbf{C}^{-1}\mathbf{W}\boldsymbol{\rho}\\
\mathbf{W}^{\text{ML}}\boldsymbol{\rho} & =\mathbf{S}\mathbf{C}^{-1}\mathbf{W}^{\text{ML}}\boldsymbol{\rho}
\end{align*}
To find the solution for the above we first express the $\mathbf{W}\boldsymbol{\rho}^{1/2}$
term using its singular value decomposition: 
\begin{equation}
\mathbf{W}\boldsymbol{\rho}^{1/2}=\mathbf{ULV}^{T}
\end{equation}
which leads to:
\begin{align*}
\mathbf{C}^{-1}\mathbf{W}\boldsymbol{\rho}^{1/2} & =\mathbf{UL}\left(\mathbf{L}^{2}+\sigma^{2}\mathbf{I}_{K}\right)^{-1}\mathbf{V}^{T}
\end{align*}
then:
\begin{align*}
\mathbf{S}\mathbf{C}^{-1}\mathbf{W}\boldsymbol{\rho}^{1/2} & =\mathbf{W}\boldsymbol{\rho}^{1/2}\\
\mathbf{S}\mathbf{UL}\left(\mathbf{L}^{2}+\sigma^{2}\mathbf{I}_{K}\right)^{-1}\mathbf{V}^{T} & =\mathbf{ULV}^{T}\\
\mathbf{S}\mathbf{UL} & =\mathbf{U}\left(\mathbf{L}^{2}+\sigma^{2}\mathbf{I}_{K}\right)\mathbf{L}
\end{align*}
which implies that $\mathbf{u}_{j}$ is the eigenvector of $\mathbf{S}$
with eigenvalue of $\lambda_{j}=\sigma^{2}+l_{j}^{2}$. Therefore
all potential solutions for $\mathbf{W}^{\text{ML}}$ may be written
as
\begin{equation}
\mathbf{W^{\text{ML}}=}\mathbf{U}_{K}\left(\mathbf{K}_{K}-\sigma^{2}\mathbf{I}_{K}\right)^{1/2}\mathbf{R}\boldsymbol{\rho}^{-1/2}
\end{equation}
where 
\begin{equation}
k_{jj}=\begin{cases}
\lambda_{j} & \mathrm{eigenvalue \ of \ } \mathbf{u}_{j}\\
\sigma^{2} & \text{otherwise}
\end{cases}
\end{equation}
where $\mathbf{R}$ is $\left(D\times K\right)$ orthonomal matrix.
The weighting term $\boldsymbol{\rho}$ allows to explicit control
over the scale of the different projection axis. $\boldsymbol{\rho}$
controls if we should place more or less importance on the role of
the input to the projection axis, which is meant to reflect our posterior
belief of re-scaling due to not all data points sharing all subspaces.
Appropriate scaling with $\boldsymbol{\rho}$ can address a well known
pitfalls of PCA such as: the disproportionate crowding of the projections
due to outliers or multi-modalities; the sphericalization of the projection

\section{Updating hyperparameters} \label{Appendix:DistributionParameters}

\subsection*{Updating $\sigma^{2}$}

We place a inverse-Gamma prior on $\sigma^{2}$ with parameters
$\left\{ \gamma,\vartheta\right\}$:
\begin{align*}
p\left(\sigma^{2}\vert\gamma,\vartheta\right) & =\frac{\vartheta^{\gamma}}{\Gamma\left(\gamma\right)}\left(\sigma^{2}\right)^{-\gamma-1}\exp\left[-\frac{\vartheta}{\left(\sigma^{2}\right)}\right]
\end{align*}
This leads to posterior distribution over $\sigma^{2}$ of the form:
\begin{align*}
p\left(\sigma^{2}\vert\gamma,\vartheta,\mathbf{Y},\mathbf{W},\mathbf{X},\mathbf{Z}\right) & =\frac{\vartheta^{\gamma}}{\Gamma\left(\gamma\right)}\left(\sigma^{2}\right)^{-\gamma-1}\exp\left[-\frac{\vartheta}{\sigma^{2}}\right]\nonumber \\
 & \times\frac{1}{\left(2\pi\sigma^{2}\right)^{\frac{ND}{2}}}\exp\left(-\frac{1}{2\sigma^{2}}\sum_{n=1}^{N}\left[\left(\mathbf{y}_{n}-\mathbf{W}\left(\mathbf{x}_{n}\odot\mathbf{z}_{n}\right)\right)^{T}\left(\mathbf{y}_{n}-\mathbf{W}\left(\mathbf{x}_{n}\odot\mathbf{z}_{n}\right)\right)\right]\right)\label{eq:SigmaPost}\\
 & \propto\left(\sigma^{2}\right)^{-\left(\gamma+ND/2\right)-1} \\
 & \times\exp\left(-\frac{1}{\sigma^{2}}\left(\frac{1}{2}\mathrm{tr}\left[\left(\mathbf{Y}-\mathbf{W}\left(\mathbf{X}\odot\mathbf{Z}\right)\right)^{T}\left(\mathbf{Y}-\mathbf{W}\left(\mathbf{X}\odot\mathbf{Z}\right)\right)\right]+\vartheta\right)\right)\nonumber 
\end{align*}
which is still a inverse-Gamma distribution with parameters $\gamma^{post}=\gamma+\frac{ND}{2}$
and $\vartheta^{post}=\frac{1}{2}\mathrm{tr}\left[\left(\mathbf{Y}-\mathbf{W}\left(\mathbf{X}\odot\mathbf{Z}\right)\right)^{T}\left(\mathbf{Y}-\mathbf{W}\left(\mathbf{X}\odot\mathbf{Z}\right)\right)\right]+\vartheta$.

\subsection*{Updating $\alpha$}

We place a Gamma prior on the IBP concentration parameter $\alpha$
with parameters $\left\{ \lambda,\mu\right\} $:
\begin{equation}
p\left(\alpha\vert\lambda,\mu\right)=\frac{\mu^{\lambda}}{\Gamma\left(\lambda\right)}\left(\alpha\right)^{\lambda-1}\exp\left[-\mu\alpha\right]
\end{equation}
This leads to posterior distribution over $\alpha$ of the form:
\begin{align}
p\left(\alpha\vert\lambda,\mu,\mathbf{Y},\mathbf{W},\mathbf{X},\mathbf{Z}\right) & =\frac{\mu^{\lambda}}{\Gamma\left(\lambda\right)}\left(\alpha\right)^{\lambda-1}\exp\left[-\mu\alpha\right]\nonumber \\
 & \times\exp\left(-\alpha H_{N}\right)\alpha^{K}\times\left(\prod_{k=1}^{K}\frac{\left(m_{k}-1\right)!\left(N-m_{k}\right)!}{\left(N\right)!}\right)\label{eq:alphaPost}\\
 & \propto\left(\alpha\right)^{\lambda+K-1}\exp\left(-\alpha\left(H_{N}+\mu\right)\right)\nonumber 
\end{align}
which is still a gamma distribution with parameters $\lambda^{post}=\lambda+K$,
$\mu^{post}=H_{N}+\mu$ and $H_{N}=\sum_{n=1}^{N}\frac{1}{n}$.

\section{Projection matrix update using \textsc{Pymanopt}} \label{Appendix:W-update}

\begin{figure}[t]
\includegraphics[scale=0.8]{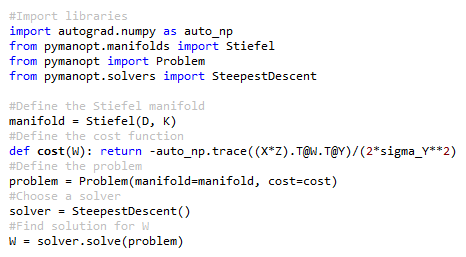}
\centering{}\caption{\label{fig:CodeUpdateW} Python code for aPPCA updates on the rotation matrix $\mathbf{W}$ using \textsc{Pymanopt} toolbox.}
\end{figure}

For both variants of the aPPCA, the matrix $\mathbf{W}$ is updated numerically by minimising the negative-log of of Equation \eqref{eq:PostW} over the Stiefel manifold with respect to the matrix $\mathbf{W}$. Figure \ref{fig:CodeUpdateW}  shows the implementation of this using the \textsc{Pymanopt} toolbox \cite{townsend2016pymanopt}.

\end{document}